%% file: main.tex
\documentclass[sigconf,screen]{acmart}
\usepackage{xspace,balance,tabularx,multirow}
\usepackage{flushend}
\usepackage{tikz}
\usepackage{pgfplots}
\pgfplotsset{compat=1.16}
\usetikzlibrary{patterns}
\usepackage{subfig}
\usepackage[ruled, vlined, linesnumbered]{algorithm2e}
\usepackage{xcolor}
\usepackage{colortbl}
\usepackage{bbold}
\SetKwComment{Comment}{$\triangleright$\ }{}
\usepackage{enumitem}
\usepackage{tablefootnote}
\usepackage{upgreek,textgreek}
\usepackage{pifont}
\usepackage[noabbrev]{cleveref}
\usepackage{titlecaps}
\usepackage{lipsum}
\usepackage{tabularx}
\usepackage{float}

\pgfplotsset{every tick label/.append style={font=\tiny}}

\newlength{\starsize}
\newlength{\starspread}
\tikzset{starsize/.code={\setlength{\starsize}{#1}},
         starspread/.code={\setlength{\starspread}{#1}}}
\tikzset{starsize=1mm,
         starspread=3mm}
\pgfdeclarepatternformonly[\starspread,\starsize]
  {my fivepointed stars}
  {\pgfpointorigin}
  {\pgfqpoint{\starspread}{\starspread}}
  {\pgfqpoint{\starspread}{\starspread}}
  {
   \pgftransformshift{\pgfqpoint{\starsize}{\starsize}}
   \pgfpathmoveto{\pgfqpointpolar{18}{\starsize}}
   \pgfpathlineto{\pgfqpointpolar{162}{\starsize}}
   \pgfpathlineto{\pgfqpointpolar{306}{\starsize}}
   \pgfpathlineto{\pgfqpointpolar{90}{\starsize}}
   \pgfpathlineto{\pgfqpointpolar{234}{\starsize}}
   \pgfpathclose%
   \pgfusepath{fill}
  }

\input{tex/macro}

\AtBeginDocument{%
  \providecommand\BibTeX{{%
    \normalfont B\kern-0.5em{\scshape i\kern-0.25em b}\kern-0.8em\TeX}}}

\copyrightyear{2024}
\acmYear{2024}
\setcopyright{rightsretained}
\acmConference[KDD '24]{Proceedings of the 30th ACM SIGKDD Conference on Knowledge Discovery and Data Mining}{August 25--29, 2024}{Barcelona, Spain}
\acmBooktitle{Proceedings of the 30th ACM SIGKDD Conference on Knowledge Discovery and Data Mining (KDD '24), August 25--29, 2024, Barcelona, Spain}
\acmDOI{XXXXXXX.XXXXXXX}
\acmISBN{978-1-4503-XXXX-X/18/06}

\begin{document}

\title{Efficient Topology-aware Data Augmentation for High-Degree Graph Neural Networks}
\subtitle{Technical Report}

\author{Yurui Lai}
\affiliation{%
  \institution{Hong Kong Baptist University}
  \country{}
}
\email{csyrlai@hkbu.edu.hk}

\author{Xiaoyang Lin}
\affiliation{%
  \institution{Hong Kong Baptist University}
  \country{}
}
\email{csxylin@hkbu.edu.hk}

\author{Renchi Yang}
\affiliation{%
  \institution{Hong Kong Baptist University}
  \country{}
}
\email{renchi@hkbu.edu.hk}

\author{Hongtao Wang}
\affiliation{%
  \institution{Hong Kong Baptist University}
  \country{}
}
\email{cshtwang@hkbu.edu.hk}

\renewcommand{\shortauthors}{Yurui Lai et al.}

\begin{abstract}
In recent years, {\em graph neural networks} (GNNs) have emerged as a potent tool for learning on graph-structured data and won fruitful successes in varied fields.
The majority of GNNs follow the {\em message-passing} paradigm, where representations of each node are learned by recursively aggregating features of its neighbors. However, this mechanism brings severe over-smoothing and efficiency issues over {\em high-degree graphs} (HDGs), wherein most nodes have dozens (or even hundreds) of neighbors, such as social networks, transaction graphs, power grids, etc. Additionally, such graphs usually encompass rich and complex structure semantics, which are hard to capture merely by feature aggregations in GNNs.

Motivated by the above limitations, we propose \algo, an efficient and effective front-mounted data augmentation framework for GNNs on HDGs. Under the hood, \algo includes two key modules: (i) feature expansion with structure embeddings, and (ii) topology- and attribute-aware graph sparsification. The former obtains augmented node features and enhanced model capacity by encoding the graph structure into high-quality structure embeddings with our highly-efficient sketching method. Further, by exploiting task-relevant features extracted from graph structures and attributes, the second module enables the accurate identification and reduction of numerous redundant/noisy edges from the input graph, thereby alleviating over-smoothing and facilitating faster feature aggregations over HDGs.
Empirically, \algo considerably improves the predictive performance of mainstream GNN models on 8 real homophilic/heterophilic HDGs in terms of node classification, while achieving efficient training and inference processes.

\end{abstract}



\maketitle

\input{tex/introduction}

\input{tex/relatedwork}

\input{tex/preliminary}

\input{tex/solution}

\input{tex/experiments}

\section{Conclusion}
In this paper, we present \algo, an efficient and effective data augmentation approach specially catered for GNNs on HDGs. \algo achieves high result utility through two main technical contributions: feature expansion with structure embeddings via hybrid sketching, and topology- and attribute-aware graph sparsification. Considerable experiments on 8 homophilic and heterophilic HDGs have verified that \algo is able to consistently promote the performance of popular MP-GNNs, e.g., GCN, GAT, SGC, APPNP, and GCNII, with matching or even upgraded training and inference efficiency. 

\begin{acks}
Renchi Yang is supported by the NSFC Young Scientists Fund (No. 62302414) and Hong Kong RGC ECS grant (No. 22202623).
\end{acks}


\clearpage
\balance
\bibliographystyle{ACM-Reference-Format}
\bibliography{main}

\appendix
\input{tex/proof}

\input{tex/other}


\input{tex/addexp}

\end{document}

%% file: tex/macro.tex
\newcommand{\renchi}[1]{{\color{red}{[Renchi: #1]}}}

\makeatletter
\newcommand*\bigcdot{\mathpalette\bigcdot@{.5}}
\newcommand*\bigcdot@[2]{\mathbin{\vcenter{\hbox{\scalebox{#2}{$\m@th#1\bullet$}}}}}
\makeatother

\newcommand{\rvline}{\hspace*{-\arraycolsep}\vline\hspace*{-\arraycolsep}}

\newcommand{\stitle}[1]{\vspace*{0.5em}\noindent{\bf #1.\/}}

\newcommand{\algo}{\textsc{TADA}\xspace}

\newcommand{\V}{\mathcal{V}\xspace}
\newcommand{\G}{\mathcal{G}\xspace}
\newcommand{\N}{\mathcal{N}\xspace}
\newcommand{\Y}{\mathcal{Y}\xspace}
\newcommand{\EDG}{\mathcal{E}\xspace}
\newcommand{\C}{\mathcal{C}\xspace}

\newcommand{\xm}{\mathbf{x}\xspace}

\newcommand{\WM}{\mathbf{W}\xspace}
\newcommand{\AM}{\mathbf{A}\xspace}
\newcommand{\NAM}{\widetilde{\mathbf{A}}\xspace}
\newcommand{\DM}{\mathbf{D}\xspace}
\newcommand{\IM}{\mathbf{I}\xspace}
\newcommand{\SM}{\mathbf{S}\xspace}
\newcommand{\MM}{\mathbf{M}\xspace}
\newcommand{\PM}{\mathbf{P}\xspace}

\newcommand{\XM}{\mathbf{X}\xspace}
\newcommand{\LM}{\mathbf{L}\xspace}
\newcommand{\UM}{\mathbf{U}\xspace}
\newcommand{\VM}{\mathbf{V}\xspace}
\newcommand{\HM}{\mathbf{H}\xspace}
\newcommand{\ZM}{\mathbf{Z}\xspace}
\newcommand{\GM}{\mathbf{G}\xspace}

\newcommand{\RM}{\mathbf{R}\xspace}

\newcommand{\LWM}{\boldsymbol{\Omega}\xspace}

\newcommand{\QM}{\mathbf{Q}\xspace}

\newenvironment{customlegend}[1][]{%
    \begingroup
    \csname pgfplots@init@cleared@structures\endcsname
    \pgfplotsset{#1}%
}{%
    \csname pgfplots@createlegend\endcsname
    \endgroup
}%

\def\addlegendimage{\csname pgfplots@addlegendimage\endcsname}

\makeatletter
\newcommand\footnoteref[1]{\protected@xdef\@thefnmark{\ref{#1}}\@footnotemark}
\makeatother

\let\oldnl\nl
\newcommand{\nonl}{\renewcommand{\nl}{\let\nl\oldnl}}

\SetKwComment{Comment}{/* }{ */}

\makeatletter 
\g@addto@macro{\@algocf@init}{\SetKwInOut{Parameter}{Parameters}} 
\makeatother

\definecolor{myred}{HTML}{fd7f6f}
\definecolor{myred_new}{HTML}{D8D8D8}
\definecolor{myred_new2}{HTML}{D7191C}
\definecolor{myblue}{HTML}{7eb0d5}
\definecolor{mygreen}{HTML}{b2e061}
\definecolor{mypurple}{HTML}{bd7ebe}
\definecolor{myorange}{HTML}{ffb55a}
\definecolor{myyellow}{HTML}{ffee65}
\definecolor{mypurple2}{HTML}{beb9db}
\definecolor{mypink}{HTML}{fdcce5}
\definecolor{mycyan}{HTML}{8bd3c7}

\definecolor{myblue2}{HTML}{115f9a}
\definecolor{myred2}{HTML}{c23728}

\newcommand{\eat}[1]{}

\definecolor{B0}{HTML}{3C2F80}
\definecolor{B1}{HTML}{012030}
\definecolor{B2}{HTML}{13678A}
\definecolor{B3}{HTML}{45C4B0}
\definecolor{B4}{HTML}{9AEBA3}
\definecolor{B5}{HTML}{DAFDBA}

\definecolor{O1}{HTML}{F29E38}
\definecolor{O2}{HTML}{F28444}

\definecolor{R1}{HTML}{F2889B}

\definecolor{my_orange}{HTML}{d95319}
\definecolor{my_green}{HTML}{117733}

%% file: tex/introduction.tex
\section{Introduction}\label{sec:intro}
{\em Graph neural networks} (GNNs) are powerful deep learning architectures for relational data (a.k.a. graphs or networks), which have exhibited superb performance in extensive domains spanning across recommender systems \cite{ying2018graph}, bioinformatics \cite{stokes2020deep,fout2017protein}, transportation \cite{derrow2021eta,jiang2022graph}, finance \cite{zhang2022efraudcom,chen2018incorporating,wang2022federatedscope}, and many other \cite{gilmer2017neural,rusek2019unveiling,lam2022graphcast,merchant2023scaling,wu2022graph}.
The remarkable success of GNN models 
is primarily attributed to
the recursive {\em message passing} (MP) (a.k.a. feature aggregation or feature propagation) scheme \cite{gilmer2017neural}, where the features of a node are iteratively updated by aggregating the features from its neighbors.

In real world, graph-structured data often encompasses a wealth of node-node connections (i.e., edges), where most nodes are adjacent to dozens or hundreds of neighbors on average, which are referred to as {\em high-degree graphs} (hereafter HDGs). Practical examples include social networks/medias (e.g., Facebook, TikTok, LinkedIn), transaction graphs (e.g., PayPal and AliPay), co-authorship networks,
airline networks, and power grids. Over such graphs, the MP mechanism undergoes two limitations: (i) homogeneous node representations after merely a few rounds of feature aggregations (i.e., over-smoothing \cite{chen2020measuring}), and (ii) considerably higher computation overhead.
Apart from this, the majority of GNNs mainly focus on designing new feature aggregation rules or model architectures, where the rich structural features of nodes in HDGs are largely overlooked and under-exploited.

To prevent overfitting and over-smoothing in GNNs, a series of studies draw inspiration from Dropout~\cite{srivastava2014dropout} and propose to randomly remove or mask edges~\cite{rong2019dropedge}, nodes \cite{feng2020graph,you2020graph}, subgraphs \cite{you2020graph} from the input graph $\G$ during model training. Although such random operations can be done efficiently, they yield information loss and sub-optimal results due to removing graph elements while overlooking their importance to $\G$ in the context of tasks.
Recently, some researchers \cite{srinivasa2020fast,liu2023dspar} applied graph sparsification techniques for better graph reduction, which is also task-unaware and fails to account for attribute information.
Instead of relying on heuristics, several attempts~\cite{zhao2021data,franceschi2019learning,jin2020graph} have been made to search for better graph structures that augment $\G$ via {\em graph structure learning}. This methodology requires expensive training, might create additional edges in $\G$, and thus can hardly cope with HDGs.
To tackle the feature-wise limitation of GNNs, recent efforts \cite{Velingker2022AffinityAwareGN,song2021topological,sun2023feature} resort to expanding node features with proximity metrics (e.g., hitting time, commute time) or network embeddings~\cite{cai2018comprehensive}, both of which are computationally demanding, especially on large HDGs.
In sum, existing augmentation techniques for GNNs either compromise effectiveness or entail significant extra computational expense when applied to HDGs.

In response, this paper proposes \algo, an effective and efficient data augmentation solution tailored for GNN models on HDGs. In particular, \algo tackles the aforementioned problems through two vital contributions: (i) efficient feature expansion with sketch-based structure embeddings; and (ii) topology- and attribute-aware graph sparsification. The former aims to extract high-quality structural features underlying the input HDG $\G$ to expand node features for bolstered model performance in a highly-efficient fashion, while the latter seeks to attenuate the adverse impacts (i.e., over-smoothing and efficiency issues) of feature aggregations on HDGs by expunging redundant/noisy edges in $\G$ with consideration of both graph topology and node attributes.

To achieve the first goal, we first empirically and theoretically substantiate the effectiveness of the intact graph structures (i.e., adjacency matrices) in improving GNNs when serving as additional node attributes. In view of its impracticality on large HDGs, we further develop a hybrid sketching approach that judiciously integrates our novel topology-aware RWR-Sketch technique into the data-oblivious Count-Sketch method for fast and accurate embeddings of graph structures. Compared to naive Count-Sketch, which offers favorable theoretical merits
but has flaws in handling highly skewed data (i.e., HDGs) due to its randomness, RWR-Sketch 
remedies this deficiency by injecting a concise summary of the HDG using the {\em random walk with restart} \cite{tong2006fast} model.
The resulted structure embeddings, together with node attributes, are subsequently transformed into task-aware node features via pre-training.
On top of that, we leverage such augmented node features for our second goal. That is, instead of direct sparsification of the HDG $\G$, we first construct an edge-reweighted graph $\G_w$ using the enriched node features. Building on our rigorous theoretical analysis, a fast algorithm for estimating the centrality values
of edges in $\G_w$ is devised for identifying unnecessary/noisy edges.

We extensively evaluate \algo along with 5 classic GNN models on 4 homophilic graphs and 4 heterophilic graphs in terms of node classification. Quantitatively, the tested GNNs generally and consistently achieve conspicuous improvements in accuracy ({up to $20.14\%$}) when working in tandem with \algo, while offering matching or superior efficiency (up to orders of magnitude speedup) in each training and inference epoch.






To summarize, our paper makes the following contributions:
\begin{itemize}[leftmargin=*]
\vspace{-1ex}
\item Methodologically, we propose a novel data augmentation framework \algo for GNNs on HDGs, comprising carefully-crafted skecthing-based feature expansion and graph sparsification.
\item Theoretically, we corroborate the effectiveness of using the adjacency matrix as auxiliary attributes and establish related theoretical bounds in our sketching and sparsification modules.
\item Empirically, we conduct experiments on 8 benchmark datasets and demonstrate the effectiveness and efficiency of \algo in augmenting 5 popular GNN models.
\end{itemize}

%% file: tex/relatedwork.tex
\vspace{-1ex}
\section{Related Works}\label{sec:related-work}
\vspace{-1ex}
\stitle{Data Augmentation for GNNs} 
Data augmentation for GNNs (GDA) aims at increasing the generalization ability of GNN models through structure modification or feature generation, which has been extensively studied in the literature~\cite{adjeisah2023towards, zhao2022graph, yang2023data, ding2022data}. Existing GDA works can be generally categorized into two types: (i) rule-based methods and (ii) learning-based methods. More specifically, rule-based GDA techniques rely on heuristics (pre-defined rules) to modify or manipulate the graph data. Similar in spirit to Dropout~\cite{srivastava2014dropout}, DropEdge~\cite{rong2019dropedge} and its variants~\cite{feng2020graph,thakoor2021large,wang2020graphcrop,Sun2021MoCLDM,fang2023dropmessage} randomly remove or mask edges, nodes, features, subgraphs, or messages so as to alleviate the over-fitting and over-smoothing issues. However, this methodology causes information loss and, hence, sub-optimal quality since the removal operations treat all graph elements equally.
In lieu of removing data, ~\citet{ying2021transformers} propose to add virtual nodes that connect to all nodes and ~\cite{wang2021mixup,han2022g,verma2021graphmix} create new data samples by either interpolating training samples~\cite{zhang2018mixup} or hidden states and labels~\cite{verma2019manifold}. Besides, recent studies \cite{liu2022local,song2021topological,Velingker2022AffinityAwareGN,sun2023feature} explored extracting additional node features from graph structures. For instance, ~\citet{song2021topological} augment node attributes with node embeddings from DeepWalk~\cite{perozzi2014deepwalk} and ~\citet{Velingker2022AffinityAwareGN} expand node features with random walk measures (e.g., effective resistance, hitting and commute times). These approaches enjoy better effectiveness at the expense of high computation costs, which are prohibitive on large HDGs.

Along another line, learning-based approaches leverage deep learning for generations of task-specific augmented samples.
Motivated by the assumption that graph data is noisy and incomplete, {\em graph structure learning} (GSL)~\cite{jin2020graph,zhao2021data,jin2022empowering} methods learn better graph structures by treating graph structures as learnable parameters. As an unsupervised learning method, {\em graph contrastive learning} (GCL) \cite{you2020graph,zhu2021graph} techniques have emerged as a promising avenue to address the challenges posed by noisy and incomplete graph data, enhancing the robustness and generalization of graph neural networks (GNNs) on high-dimensional graphs (HDGs).
Unlike GSL and GCL, \cite{kong2020flag,jin2019latent,xu2019topology} extend adversarial training to graph domains and augments input graphs with adversarial patterns by perturbing node features or graph structures during model training. Rationalization methods~\cite{liu2022graph,wu2021discovering} seek to learn subgraphs that are causally related with the graph labels as a form of augmented graph data, which are effective in solving out-of-distribution and data bias issues.
Recently, researchers \cite{zhao2022autogda,luo2022automated,you2021graph} utilized reinforcement learning agents to automatically learn optimal augmentation strategies for different subgraphs or graphs. These learning-based approaches are all immensely expensive, and none of them tackle the issues of GNNs on HDGs as remarked in Section~\ref{sec:intro}.

\stitle{Structure Embedding} The goal of structure embedding (or network embedding) is to convert the graph topology surrounding each node into a low-dimensional feature vector. As surveyed in \cite{cai2018comprehensive}, there exists a large body of literature on this topic, most of which can be summarized into three categories as per their adopted methodology: (i) random walk-based methods, (ii) matrix factorization-based methods, and (iii) deep learning-based models. In particular, random walk-based methods \cite{perozzi2014deepwalk,grover2016node2vec,tsitsulin2018verse} learn node embeddings by optimizing the skip-gram model~\cite{mikolov2013efficient} or its variants with random walk samples from the graph. Matrix factorization-based approaches~\cite{qiu2018network,yang2020homogeneous,ou2016asymmetric,zhang2018arbitrary,yang2022scalable} construct node embeddings through factorizing node-to-node affinity matrices, whereas \cite{wang2016structural,cao2016deep,abu2018watch,yu2018learning} capitalize on diverse deep neural network models for node representation learning on non-attributed graphs. Recent evidence suggests that using such network embeddings~\cite{song2021topological}, or resistive embeddings~\cite{Velingker2022AffinityAwareGN} and spectral embeddings~\cite{sun2023feature} as complementary node features can bolster the performance of GNNs, but result in considerable additional computational costs.

\stitle{Graph Sparsification} Graph sparsification is a technique aimed at approximating a given graph $\G$ with a sparse graph containing a subset of nodes and/or edges from $\G$~\cite{chen2023demystifying}. Classic sparsification algorithms for graphs include cut sparsification~\cite{karger1994random,fung2011general} and spectral sparsification~\cite{spielman2008graph,batson2009twice,spielman2004nearly}.
Cut sparsification reduces edges while preserving the value of the graph cut, while spectral sparsifiers ensure the sparse graphs can retain the spectral properties of the original ones.
Recent studies~\cite{srinivasa2020fast,liu2023dspar} employ these techniques as heuristics to sparsify the input graphs before feeding them into GNN models for acceleration of GNN training~\cite{liu2022survey}.
In spite of their improved empirical efficiency, these works fail to incorporate node attributes as well as task information for sparsification.
To remove task-irrelevant edges accurately, \citet{zheng2020robust} and \citet{li2020sgcn} cast graph sparsification as optimization problems and apply deep neural networks and the alternating direction method of multipliers, respectively, both of which are cumbersome for large HDGs. 

%% file: tex/preliminary.tex
\section{Preliminaries}

\subsection{Notations}
Throughout this paper, sets are denoted by calligraphic letters, e.g., $\V$. Matrices (resp. vectors) are written in bold uppercase (resp. lowercase) letters, e.g., $\MM$ (resp. $\mathbf{x}$). We use $\MM_{i}$ and $\MM_{:,i}$ to represent the $i^{\textnormal{th}}$ row and column of $\MM$, respectively.

Let $\G=(\V,\EDG)$ be a graph (a.k.a. network), where $\V$ is a set of $n$ nodes and $\EDG$ is a set of $m$ edges. For each edge $e_{i,j}\in \EDG$ connecting nodes $v_i$ and $v_j$, we say $v_i$ and $v_j$ are neighbors to each other. We use $\N(v_i)$ to denote the set of neighbors of node $v_i$, where the degree of $v_i$ (i.e., $|\N(v_i)|$) is symbolized by $d(v_i)$. Nodes $v_i$ in $\G$ are endowed with an attribute matrix $\XM\in \mathbb{R}^{n\times d}$, where $d$ stands for the dimension of node attribute vectors. The diagonal degree matrix of $\G$ is denoted as $\DM=\textsf{diag}(d(v_1),\cdots,d(v_n))$. The adjacency matrix and normalized adjacency matrix are denoted as $\AM$ and $\NAM = \DM^{-\frac{1}{2}}\AM\DM^{-\frac{1}{2}}$, respectively. The Laplacian and transition matrices of $\G$ are defined by $\LM=\DM-\AM$ and $\PM=\DM^{-1}\AM$, respectively.

\subsection{Graph Neural Networks (GNNs)}
The majority of existing GNNs \cite{gasteiger2019diffusion,wu2019simplifying,velivckovic2018graph,xu2018powerful,bianchi2021graph,defferrard2016convolutional,xu2018representation,chien2020adaptive,liu2020towards,huang2023node} follow the {\em message passing} (MP) paradigm \cite{gilmer2017neural}, such as GCN \cite{kipf2016semi}, APPNP \cite{gasteiger2018predict}, and GCNII \cite{chen2020simple}. For simplicity, we refer to all these MP-based models as GNNs. More concretely, the node representations $\HM^{(t)}$ at $t$-th layer of GNNs can be written as
\begin{equation}
\begin{gathered}
\HM^{(t)} = \sigma(f_{\text{trans}}(f_{\text{aggr}}(\G,\HM^{(t-1)}))),\\
\HM^{(0)}=\sigma(\XM\LWM_{\textnormal{orig}})\in \mathbb{R}^{n\times h}
\end{gathered}
\end{equation}
where $\sigma(\cdot)$ stands for a nonlinear activate function, $f_{\text{trans}}(\cdot)$ corresponds to a layer-wise feature transformation operation (usually an MLP including non-linear activation ReLU and layer-specific learnable weight matrix), and $f_{\text{aggr}}(\G,\cdot)$ represents the operation of aggregating $\ell$-th layer features $\HM^{(\ell)}$ from the neighborhood along graph $\G$, e.g., $f_{\text{aggr}}(\G,\HM^{(t)})=\NAM\HM^{(t-1)}$ in GCN and $f_{\text{aggr}}(\G,\HM^{(t)})=(1-\alpha)\NAM\HM^{(t-1)}+\alpha\cdot \HM^{(0)}$ in APPNP.
Note that $\HM^{(0)}=\sigma(\XM\LWM_{\textnormal{orig}})\in \mathbb{R}^{n\times h}$ is the initial node features resulted from a non-linear transformation from the original node attribute matrix $\XM$ using an MLP parameterized by learnable weight $\LWM_{\textnormal{orig}}$. As demystified in a number of studies~\cite{sun2023feature,zhu2021interpreting,gasteiger2019diffusion,wang2021approximate,wu2019simplifying}, after removing non-linearity, the node representations $\HM^{(t)}$ learned at the $t$-layer in most MP-GNNs can be rewritten as linear approximation formulas:
\begin{equation}\label{eq:poly}
\HM^{(t)} = f_{\textnormal{poly}}(\NAM,t)\cdot\XM\cdot\LWM,
\end{equation}
where $f_{\textnormal{poly}}(\NAM,t)$ stands for a $t$-order polynomial, $\NAM$ (or $\PM$) is the structure matrix of $\G$, and $\LWM$ is the learned weight. For instance, $f_{\textnormal{poly}}(\GM,t)=\NAM^{t}\XM$ in GCN and $f_{\textnormal{poly}}(\GM,t)=\sum_{i=0}^{t}{\alpha^i\NAM^i}\XM$ in APPNP. 

\subsection{GNNs over High-Degree Graphs (HDGs)}\label{sec:HDG-limitations}

\begin{table}[!t]
\centering
\renewcommand{\arraystretch}{0.9}
\begin{small}
\caption{Classification Accuracy with $\XM\mathbin\Vert\AM$ as Features.}\label{tbl:acc}
\vspace{-3mm}
\begin{tabular}{l|cc}
	\hline
	{\bf Dataset}  &  \multicolumn{1}{c}{\bf WikiCS } & \multicolumn{1}{c}{\bf Squirrel  }  \\
	\hline
    \hline
    GCN  & $84.05\%_{\pm 0.76\%}$  &   $54.85\%_{\pm 2.02\%}$ \\
    GCN ($\XM\mathbin\Vert\AM$)   & $\bf 84.15\%_{\pm 0.48\%}$  & $\bf 56.06\%_{\pm 3.12\%}$    \\
    \hline
    GAT & $83.74\%_{\pm 0.75\%}$  &  $55.70\%_{\pm 3.26\%}$   \\
    GAT ($\XM \mathbin\Vert \AM$) & $\bf 84.15\%_{\pm 0.83\%}$ &  $\bf 57.77\%_{\pm 2.13\%}$   \\
    \hline
    APPNP & $85.04\%_{\pm 0.60\%}$  & $54.47\%_{\pm 2.06\%}$  \\
    APPNP ($\XM \mathbin\Vert \AM$) & $\bf 85.24\%_{\pm 0.56\%}$ & $\bf 57.64\%_{\pm 1.32\%}$     \\
    \hline
    GCNII   & $85.13\%_{\pm 0.56\%}$ & $ 53.13\%_{\pm 4.29\%}$  \\
    GCNII ($\XM \mathbin\Vert \AM$)  & $\bf 85.28\%_{\pm 0.78\%}$ & $\bf 54.35\%_{\pm 4.04\%}$    \\
    \hline
\end{tabular}%
\vspace{-2mm}
\end{small}
\end{table}




Although GNNs achieve superb performance by the virtue of the {\em feature aggregation} mechanism, they incur severe inherent drawbacks, which are exacerbated over HDGs, as analysed below.

\stitle{Inadequate Structure Features}
Intuitively, structure features play more important roles for HDGs as they usually encompass rich and complex topology semantics. However, the extant GNNs primarily capitalize on the graph structure for feature aggregation, failing to extract the abundant topology semantics underlying $\G$.
\eat{Another recent study \cite{sun2023feature} pinpoint that features will correlate and propose to add structure embeddings or positional encoding to increase feature spaces.}


To validate this observation, we conduct a preliminary empirical study with 4 representative GNN models on 2 benchmarking HDGs \cite{pei2019geom,mernyei2020wiki} in terms of node classification. 
Table \ref{tbl:acc} manifests that by concatenating the input attribute matrix $\XM$ (i.e., $\XM\mathbin\Vert\AM$) with the adjacency matrix $\AM$ as node features, each GNN model can see performance gains ({up to $3.17\%$}). 
In Appendix \ref{sec:A-Feat}, we further theoretically show that expanding features with $\AM$ can alleviate the {\em feature correlation} \cite{sun2023feature} in standard GNNs and additionally incorporate high-order proximity information between nodes as in traditional {\em network embedding} techniques \cite{grover2016node2vec,qiu2018network}. 

However, this simple trick demands learning an $(n+d)\times h$ transformation weight matrix $\LWM_{\textnormal{orig}}$, and hence, leads to the significant expense of training.

\stitle{Over-Smoothing}
Note that HDGs often demonstrate high connectivity, i.e., nodes are connected to dozens or even hundreds of neighbors on average, which implies large spectral gaps $\lambda$ \cite{chung1997spectral}.
The matrix powers $\NAM^t$ and $\PM^t$ will hence quickly converge to stationary distributions as $t$ increases, as indicated by Theorem \ref{eq:smooth}.
\begin{theorem}\label{eq:smooth}
Suppose that $\G$ is a connected and non-bipartite graph. Then, we have
\begin{small}
\begin{gather*}
\left|\NAM^t_{i,j}-\frac{\sqrt{d(v_j)\cdot d(v_i)}}{2m}\right|\le \left(1-\lambda\right)^t,\ \left|\PM^t_{i,j}-\frac{d(v_j)}{2m}\right|\le \sqrt{\frac{d(v_j)}{d(v_i)}}\left(1-\lambda\right)^t,
\end{gather*}
\end{small}
where $\lambda<1$ is the spectral gap of $\G$.
\begin{proof}
Let $\sigma_1\ge \sigma_2\ge \cdots \ge \sigma_n$ be the eigenvalues of $\NAM$ and $\sigma$ is defined by $\sigma=\min\{|\sigma_2|,|\sigma_n|\}$. By definition, the spectral gap of $\G$ is then $\lambda=1-\sigma$.
By Theorem 5.1 in \cite{lovasz1993random}, it is straightforward to get
\begin{equation}\label{eq:Ptij}
\left|\PM^t_{i,j}-\frac{d(v_j)}{2m}\right|\le \sqrt{\frac{d(v_j)}{d(v_i)}}\left(1-\lambda\right)^t.
\end{equation}
Recall that $\NAM=\DM^{-\frac{1}{2}}\AM\DM^{-\frac{1}{2}}$ and $\PM=\DM^{-1}\AM$. Hence,
\begin{equation*}
\NAM^t = \DM^{-\frac{1}{2}}\AM(\DM^{-1}\AM)^{t-1}\DM^{-\frac{1}{2}}=\DM^{\frac{1}{2}}\PM^{t}\DM^{-\frac{1}{2}},
\end{equation*}
meaning that $\NAM^t_{i,j} = \sqrt{\frac{d(v_i)}{d(v_j)}}\cdot\PM^t_{i,j}$. Plugging this into Eq. \eqref{eq:Ptij} yields
\begin{align*}
&\left|\sqrt{\frac{d(v_i)}{d(v_j)}}\PM^t_{i,j}-\sqrt{\frac{d(v_i)}{d(v_j)}}\cdot\frac{d(v_j)}{2m}\right|\le \sqrt{\frac{d(v_i)}{d(v_j)}}\sqrt{\frac{d(v_j)}{d(v_i)}}\left(1-\lambda\right)^t\\
&=\left|\NAM^t_{i,j}-\frac{\sqrt{d(v_j)\cdot d(v_i)}}{2m}\right|\le \left(1-\lambda\right)^t.
\end{align*}
which finishes the proof.
\end{proof}
\end{theorem}
As an aftermath, for any node $v_i\in V$, its node representation $\HM^{(t)}_i$ obtained in Eq. \eqref{eq:poly} turns into
\begin{gather*}
\sqrt{d(v_i)}\sum_{v_j\in \V}{\frac{\sqrt{d(v_j)}}{2m}\sum_{\ell=1}^{d}\XM_{j}\LWM}\ \text{and}\ 
\sum_{v_j\in \V}{\frac{\sqrt{d(v_j)}}{2m}\XM_{j}\LWM}
\end{gather*}
when $f_{\textnormal{poly}}(\NAM,t)=\NAM^t$ and $\PM^t$, respectively, both of which are essentially irrelevant to node $v_i$. In other words, the eventual representations of all nodes are overly smoothed with high homogeneity, rendering nodes in different classes indistinguishable and resulting in degraded model performance \cite{chen2020measuring,chen2020simple,cai2020note}.
\eat{
Figure \ref{} compares the homogeneity of node representations on normal graphs with those on HDGs when varying the number $t$ of layers of feature aggregations. \renchi{Need a figure containing data from real dataset and its dense version.}
}

\stitle{Costly Feature Aggregation} Aside from the over-smoothing, the sheer amount of feature aggregation operations of GNNs over HDGs, especially on sizable ones, engender vast computation cost. Recall that each round of feature aggregation in GNNs consumes $O(mh)$ time. Compared to normal scale-free graphs with average node degrees $m/n=O(\log(n))$ or smaller, the average node degrees in HDGs can be up to hundreds, which are approximately $O(\log^2(n))$. This implies an $O(n\log^2(n)\cdot h)$ asymptotic cost in total for each round of feature aggregation.

A workaround to mitigate the over-smoothing and computation issues caused by the feature aggregation on HDGs is to sparsify $\G$ by identifying and eradicating unnecessary or redundant edges. However, the accurate and efficient identification of such edges for improving GNN models is non-trivial in the presence of node attributes and labels and remains under-explored.

In sum, we need to address two technical challenges:
\begin{itemize}[leftmargin=*]
\item How to encode $\AM$ into high-quality structure embeddings that can augment GNNs for better model capacity on HDGs in an efficient manner?
\item How to sparsify the input HDG so as to enable faster feature aggregation while retaining the predictive power?
\end{itemize}

%% file: tex/solution.tex
\section{Methodology}
This section presents our \algo framework for tackling the foregoing challenges. Section \ref{sec:overview} provides an overview of \algo, followed by detailing its two key modules in  Sections \ref{sec:sketched-embedding} and \ref{sec:sparsification}, respectively.

\begin{figure}[!t]
\centering
\includegraphics[width=\columnwidth]{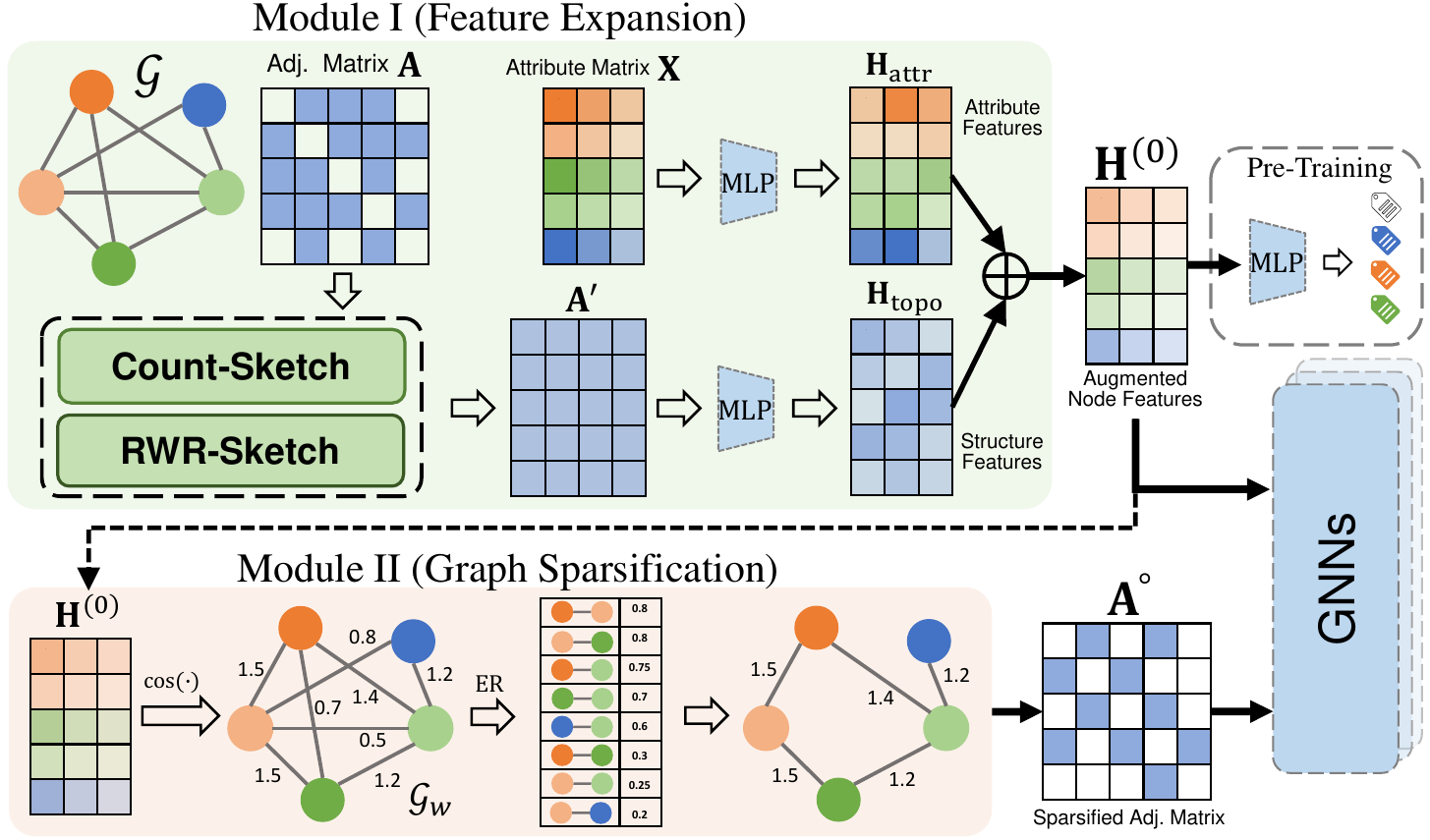}
\vspace{-5ex}
\caption{Overview of \algo}
\label{fig:overview}
\vspace{-2ex}
\end{figure}

\subsection{Synoptic Overview of \algo}\label{sec:overview}
As illustrated in Figure \ref{fig:overview}, \algo acts as a front-mounted stage for MP-GNNs, which compromises two main ingredients: (i) {\em feature expansion with structure embeddings} (\textbf{Module I}), and (ii) {\em topology- and attribute-aware graph sparsification} (\textbf{Module II}). The goal of the former component is to generate high-quality structure embeddings $\HM_{\textnormal{topo}}$ capturing the rich topology semantics underlying $\G$ for feature expansion, while the latter aims to sparsify the structure of the input graph $\G$ so as to eliminate redundant or noisy topological connections in $\G$ with consideration of graph topology and node attributes.

\stitle{Module I: Feature Expansion}
To be more specific, in Module I, \algo first applies a hybrid sketching technique (Count-Sketch + RWR-Sketch) to the adjacency matrix $\AM$ of $\G$ and transforms the sketched matrix $\AM^{\prime}\in \mathbb{R}^{n\times k}$ ($k\ll n$, typically $k=128$) into the structure embeddings $\HM_{\textnormal{topo}}\in \mathbb{R}^{n\times h}$ of all nodes via an MLP:
\begin{equation}\label{eq:H-topo}
\HM_{\textnormal{topo}} = \sigma(\AM^{\prime}\LWM_{\textnormal{topo}}),
\end{equation}
where $\sigma(\cdot)$ is a non-linear activation function (e.g., ReLU) and $\LWM_{\textnormal{topo}}\in \mathbb{R}^{k\times h}$ stands for learnable transformation weights. 
In the meantime, Module I feeds the node attribute matrix $\XM$ to an MLP network parameterized by learnable weight $\LWM_{\textnormal{attr}}\in \mathbb{R}^{d\times h}$ to obtain the transformed node attribute features $\HM_{\textnormal{attr}}\in \mathbb{R}^{n\times h}$.
\begin{equation}\label{eq:H-attr}
\HM_{\textnormal{attr}} = \sigma(\XM\LWM_{\textnormal{attr}})
\end{equation}

A linear combination of the structure embeddings $\HM_{\textnormal{topo}}$ and transformed node attribute features $\HM_{\textnormal{attr}}$ as in Eq. \eqref{eq:init-feat} yields the initial node representations $\HM^{(0)}$.
\begin{equation}\label{eq:init-feat}
\HM^{(0)} = (1-\gamma)\cdot\HM_{\textnormal{attr}} + \gamma\cdot\HM_{\textnormal{topo}}  \in \mathbb{R}^{n\times h}
\end{equation}
The hyper-parameter $\gamma$ controls the importance of node topology in the resulting node representations. 

Notice that $\HM^{(0)}$ and related learnable weights are {\em pre-trained} by the task (i.e., node classification) with a single-layer MLP as classifier (using $n_p$ epochs).
In doing so, we can extract task-specific features in $\HM^{(0)}$ to facilitate the design of Module II, and all these intermediates can be reused for subsequent GNN training.

\stitle{Module II: Graph Sparsification}
Since $\HM^{(0)}$ captures the task-aware structure and attribute features of nodes in $\G$, 
Module II can harness it to calculate the centrality values of all edges that assess their importance to $\G$ in the context of node classification.
Given the sparsification raio $\rho$, the edges with $m\cdot\rho$ lowest centrality values are therefore removed from $\G$, whereas those important ones will be kept and reweighted by the similarities of their respective endpoints in $\HM^{(0)}$. Based thereon, \algo creates a sparsified adjacency matrix denoted as $\AM^{\circ}$ as a substitute of $\AM$.
The behind intuition is that adjacent nodes with 
low connectivity and attribute homogeneity are more likely to fall under disparate classes, and hence, their direct connection (i.e., edges) can be removed without side effects.


Finally, the augmented initial node representations $\HM^{(0)}$ and the sparsified adjacency matrix $\AM^{\circ}$ are input into the MP-GNN models $f_{\textsf{GNN}}(\cdot,\cdot)$ for learning final node representations:
\begin{equation*}
\HM = f_{\textsf{GNN}}(\HM^{(0)}, \AM^{\circ})
\end{equation*}
and performing the downstream task, i.e., node classification.

In the succeeding subsections, we elaborate on the designs and details of Module I and Module II. 

\subsection{Efficient Feature Expansion with Structure Embeddings}\label{sec:sketched-embedding}
Recall that in Module I, the linchpin to the feature expansion (i.e., building structure embeddings $\HM_{\textnormal{topo}}$) is $\AM^{\prime}\in \mathbb{R}^{n\times k}$, a sketch of the adjacency matrix $\AM$. Notice that even for HDGs, $\AM$ is highly sparse ($m\ll n^2$) and the distribution of node degrees (i.e., the numbers of non-zero entries in rows/columns) is heavily skewed, rendering existing sketching tools for dense matrices unsuitable.
In what follows, we delineate our hybrid sketching approach specially catered for adjacency matrix $\AM$.



\stitle{Count-Sketch Method}
To deal with the sparsity of $\AM$, our first-cut solution is the count-sketch (or called sparse embedding) \cite{clarkson2013low} technique, which achieves $O(\textnormal{nnz}(\AM))=O(m)$ time for computing the sketched adjacency matrix $\AM^{\prime}\in \mathbb{R}^{n\times k}$: 
\begin{equation}\label{eq:sketch-A}
\AM^{\prime}=\AM\RM^\top\ \text{where}\ \RM=\boldsymbol{\Phi}\boldsymbol{\Delta}
\end{equation}
The {\em count-sketch matrix} (a.k.a. sparse embedding) $\RM\in \mathbb{R}^{k \times n}$ is randomly constructed by 
where 
\begin{itemize}[leftmargin=*]
\item $\boldsymbol{\Delta}\in\mathbb{R}^{n\times n}$ is a diagonal matrix with each diagonal entry independently chosen to be $1$ or $-1$ with probability $0.5$, and
\item $\boldsymbol{\Phi}\in \{0,1\}^{k\times n}$ is a binary matrix with $\boldsymbol{\Phi}_{h(i),i}=1$ and $0$ otherwise $\forall{1\le i\le n}$. The function $h(\cdot)$ maps $i$ ($1\le i\le n$) to $h(i) = j \in \{1,2,\cdots,k\}$ uniformly at random.
\end{itemize} 
\eat{
$\mathbb{E}[\boldsymbol{\Phi}\boldsymbol{\Phi}^{\top}]=\frac{n}{d}\cdot \IM\in \mathbb{R}^{d\times d}$ and $\boldsymbol{\Delta}^{\top}\boldsymbol{\Delta}=\boldsymbol{\Delta}\boldsymbol{\Delta}^{\top}=\IM$. So, $\mathbb{E}[\RM\RM^{\top}]=\mathbb{E}[\boldsymbol{\Phi}\boldsymbol{\Delta}^{\top}\boldsymbol{\Delta}\boldsymbol{\Phi}^{\top}]=\frac{n}{d}$.
}


In Theorem \ref{lem:ARRW}, we prove that the count-sketch matrix $\RM$ is able to create an accurate estimator for the product of $\AM$ and any matrix $\WM$ with rigorous accuracy guarantees. 
\begin{theorem}\label{lem:ARRW}
Given any matrix $\WM$ with $n$ rows and a count-sketch matrix $\RM\in  \mathbb{R}^{k\times n}$ with $k=\frac{2 \max_{v_i\in \V}{d(v_i)}}{\epsilon^2\delta}\cdot \max_{j}{\|\WM_{:,j}\|^{2}_{2}}$, the following inequality
$$
\mathbb{P}[ | (\AM_i\RM^{\top})\cdot (\RM\WM_{:,j}) - \AM_{i}\WM_{:,j} | < \epsilon ] > 1 - \delta 
$$
holds for any node $v_i\in \V$ and $j\in [1,k]$. 
\begin{proof}
Let $ \XM = (\AM\RM^{\top})\cdot  (\RM \WM) - \AM\WM$. 
Then $\XM_{i,j} = (\AM_{i}\RM^{\top})\cdot (\RM  \WM_{:,j}) - \AM_{i}\WM_{:,j} $, where $\AM_i$ is the $i$-th row of $\AM$ and $\WM_{:,j}$ is the $j$-th column of $\WM$. 
According to Lemma 4.1 in \cite{pham2014power}, for any two column vectors $\mathbf{x},\mathbf{y}\in \mathbb{R}^{n}$, $\mathbb{E}[(\RM\mathbf{x})^{\top}\cdot (\RM\mathbf{y})]=\mathbf{x}^{\top}\cdot \mathbf{y}$. Then, we have
$$
\mathbb{E}(\XM_{i,j}) \\
=  \mathbb{E}\left( (\RM(\AM_{i})^{\top})^{\top}\cdot (\RM\WM_{:,j})\right) - \AM_{i}\WM_{:,j}  \\
= 0
$$
Thus, for $1\le i,j\le n$, $(\AM_{i}\RM^{\top})\cdot (\RM  \WM_{:,j})$ is an unbiased estimator of $\AM_{i}\WM_{:,j}$.

Moreover, by Lemma 4.2 in \cite{pham2014power} and the Cauchy–Schwarz inequality, we have
\begin{align*}
Var(\XM_{i,j}) & \le \frac{1}{k}\cdot \left( (\AM_{i}\WM_{:,j})^2 + \|\AM_i\|^{2}_{2}\cdot \|\WM_{:,j}\|^{2}_{2} \right)\\
& \le \frac{2}{k}\cdot \left( \|\AM_i\|^{2}_{2}\cdot \|\WM_{:,j}\|^{2}_{2} \right) = \frac{2 d(v_i)}{k}\cdot \|\WM_{:,j}\|^{2}_{2}. 
\end{align*}
We have $\mathbb{E}( \XM^{2}_{i,j}) = Var(\XM_{i,j}) - \mathbb{E}( \XM_{i,j})^{2} \le \frac{2 d(v_i)}{k}\cdot \|\WM_{:,j}\|^{2}_{2}$.
Using Chebyshev’s Inequality, we have 

$$
\mathbb{P}[ \XM^{2}_{i,j} \ge \epsilon  ] \le  \frac{\mathbb{E}( \XM^{2}_{i,j})}{\epsilon^2} \le \frac{\frac{2 d(v_i)}{k}\cdot \|\WM_{:,j}\|^{2}_{2}}{\epsilon^2}
$$
By setting $k=\frac{2\max_{v_i\in \V}d(v_i)}{\epsilon^2\delta}\cdot \max_{j}{\|\WM_{:,j}\|^{2}_{2}}$, we can guarantee
\begin{equation*}
\mathbb{P}[|(\AM_{i}\RM^{\top})\cdot (\RM  \WM_{:,j}) - \AM_{i}\WM_{:,j}| < \epsilon] > 1-\delta,
\end{equation*}
which completes the proof.
\end{proof} 
\end{theorem}

Recall that the ideal structure embeddings $\HM^{\ast}_{\textnormal{topo}}$ is obtained when $\AM^{\prime}$ is replaced by the original adjacency matrix $\AM$ in Eq. \eqref{eq:H-topo}, i.e., $\HM^{\ast}_{\textnormal{topo}}=\sigma(\AM\LWM_{\textnormal{topo}})$.
Assume that $\WM=\LWM_{\textnormal{topo}}$ is the learned weights in this case.
If we input $\AM^{\prime}=\AM\RM^{\top}$ to Eq. \eqref{eq:H-topo} and assume the newly learned weight matrix is $\LWM_{\textnormal{topo}}=\RM\WM$, the resulted structure embeddings $\HM_{\textnormal{topo}}$ will be similar to the ideal one $\HM^{\ast}_{\textnormal{topo}}$ according to Theorem \ref{lem:ARRW}, establishing a theoretical assurance for deriving high-quality structure embeddings $\HM_{\textnormal{topo}}$ from $\AM^{\prime}$.

{
By Theorem \ref{lem:ARRW}, we can further derive the following properties of $\AM^{\prime}$ in preserving the structure in $\G$:
\begin{itemize}[leftmargin=*]
\item {\bf Property 1:} For any two nodes $v_i,v_j\in \V$, $\AM^{\prime}_{i}\cdot {\AM^{\prime}_j}^{\top}$ is an approximation of the number of common neighbors $|\N(v_i)\cap \N(v_j)|$ . Particularly, $\|\AM^{\prime}_i\|^2_2$ approximates the degree $d(v_i)$ of node $v_i$.
\item {\bf Property 2:} For any two nodes $v_i,v_j\in \V$, $(\AM^{\prime}{\AM^{\prime}}^{\top})^{t}$ is an approximation of high-order proximity matrix $\AM^{2t}$, where each $(i,j)$-th entry denotes the number of length-$2t$ paths between nodes $v_i$ and $v_j$.
\item {\bf Property 3:} Let $\widehat{\AM^{\prime}}$ be the row-based $L_2$ normalization of $\AM^{\prime}$. For any two nodes $v_i,v_j\in \V$, $\widehat{\AM^{\prime}}\widehat{\AM^{\prime}}^{\top}$ is an approximation of $(\PM\PM^{\top})^{t}$, where each $(i,j)$-th entry denotes the probability of two length-$t$ random walks originating from $v_i$ and $v_j$ meeting at any node.
\end{itemize}
Due to the space limit, we defer the proofs to Appendix~\ref{sec:proof-of-add-properties}.
}

\eat{
We can further derive additional properties of $\AM^{\prime}$ in preserving the structure in $\AM$ based on Theorem \ref{lem:ARRW}. Due to the space limit, we defer them to Appendix~\ref{sec:module-1-add}.
}






\stitle{Limitation of Count-Sketch}
Despite the theoretical merits of approximation guarantees and high efficiency offered by the count-sketch-based approach, it is data-oblivious (i.e., the sketching matrix is randomly generated) and is likely to produce poor results, especially in dealing with highly skewed data (e.g., adjacency matrices).
To explain, we first interpret $\boldsymbol{\Phi}$ as a randomized clustering membership indicator matrix, where $\boldsymbol{\Phi}_{h(i),i}=1$ indicates assigning each node $v_i$ to $h(i)$-th ($h(i)\in \{1,\cdots,k\}$) cluster uniformly at random. Each diagonal entry in $\boldsymbol{\Delta}$ is either $1$ or $-1$, which signifies that the cluster assignment in $\boldsymbol{\Phi}$ is true or false. As such, each entry $\RM_{i,j}$ represents
\begin{equation}
 \RM_{j,i} =   \begin{cases}
 \ 1\quad \text{node $v_i$ belongs to the $j$-th cluster}\\
 -1\quad \text{node $v_i$ does not belong to the $j$-th cluster}\\
 \ 0\quad \text{otherwise}.
    \end{cases}
\end{equation}
Accordingly, $\AM^{\prime}_{i,j}$ quantifies the strength of connections from $v_i$ to the $j$-th cluster via its neighbors. Since $\boldsymbol{\Phi}$ is randomly generated, distant (resp. close) nodes might fall into the same (resp. different) clusters, resulting in a distorted distribution in $\AM^{\prime}$.

\stitle{Optimization via RWR-Sketch}
As a remedy, we propose RWR-Sketch to create a structure-aware sketching matrix $\SM\in \mathbb{R}^{k\times n}$. \algo will combine $\SM$ with count sketch matrix $\RM$ to obtain the final sketched adjacency matrix $\AM^{\prime}$:
\begin{equation}\label{eq:A-prime}
\AM^{\prime} = \AM \cdot (\RM^{\top}+\beta\cdot \SM^{\top}),
\end{equation}
where $\beta$ is a hyper-parameter controlling the contribution of the RWR-Sketch in the result.
Unlike $\RM$, the construction of $\SM$ is framed as clustering $n$ nodes in $\G$ into $k$ disjoint clusters as per their topological connections to each other in $\G$. Here, we adopt the prominent {\em random walk with restart} (RWR) model \cite{tong2006fast,yang2021effective} to summarize the multi-hop connectivity between nodes. To be specific, we construct $\SM$ as follows:
\begin{itemize}
\item[(i)] We select a set $\C$ of nodes ($k\le |\C|\ll n$) with highest in-degrees from $\V$ as the candidate cluster centroids.
\item[(ii)] For each node $v_i\in \V$, we compute the RWR score $\pi(v_i,v_j)$ of every node $v_j$ in $\C$ w.r.t. $v_i$ through $T$ {\em power iterations}:
\begin{equation}
 \pi(v_i,v_j) = \sum_{t=0}^{T}{(1-\alpha)\alpha^t\PM}_{i,j},
\end{equation}
where $\alpha\in (0,1)$ is a decay factor ($0.5$ by default).
\item[(iii)] Denote by $\pi(v_j)=\sum_{v_i\in \V}{\frac{\pi(v_i,v_j)}{n}}$ the centrality (i.e., PageRank \cite{page1998pagerank}) of $v_j\in \C$. We select a set $\C_k$ of $k$ nodes from $\C$ with the largest centralities as the final cluster centroids.
\item[(iv)] For each node ${v_i\in \V}$, we pick the node $v_j\in \C_k$ with the largest RWR score $\pi(v_i,v_j)$ as its cluster centroid and set $\SM_{j,i}=1$. 
\item[(v)] After that, we give $\SM$ a final touch by applying an $L_2$ normalization for each row.
\end{itemize}

For the interest of space, we refer interested readers to Appendix \ref{sec:module-1-add} for the complete pseudo-code and detailed asymptotic analysis of our hybrid sketching approach.



\subsection{Topology- and Attribute-Aware Graph Sparsification}\label{sec:sparsification}

\stitle{Edge Reweighting}
With the augmented initial node features $\HM^{(0)}$ (Eq. \eqref{eq:init-feat}) at hand, for each edge $e_{i,j}\in \EDG$, we assign the cosine similarity of the representations of its endpoints $v_i$ and $v_j$ as the weight of $e_{i,j}$:
\begin{equation}\label{eq:e-weight}
w(e_{i,j}) = \textnormal{cos}\left(\HM^{(0)}_i,\HM^{(0)}_j\right).
\end{equation}
Accordingly, the ``degree'' of node $v_i$ can be calculated via Eq. \eqref{eq:d-weight}, which is the sum of weights of edges incident to $v_i$.
\begin{equation}\label{eq:d-weight}
d_w(v_i)=\sum_{v_j\in \N(v_i)}{w(e_{i,j})}
\end{equation}
Denote by $\G_w=(\V,\EDG_w)$ this edge-reweighted graph. The subsequent task is hence to sparsify $\G_w$.

In the literature, a canonical methodology \cite{spielman2008graph} to create the sparsified graph $\G^{\prime}$
is to sample edges with probability proportional to their {\em effective resistance} (ER) \cite{lovasz1993random} values and add them with adjusted weights to $\G^{\prime}$.
Theoretically, $\G^{\prime}$ is an unbiased estimation of the original graph $\G$ in terms of the graph Laplacian \cite{spielman2008graph} and requires $n_r=O\left(\frac{n\log(n/\delta)}{\epsilon^2}\right)$ samples to ensure the Laplacian matrix $\LM^{\prime}$ of $\G^{\prime}$ satisfies
\begin{equation*}
\forall{\xm\in \mathbb{R}^n}, \epsilon\in (0,1]\quad (1-\epsilon)\xm^{\top}\LM\xm \le \xm^{\top}\LM^{\prime}\xm \le (1+\epsilon)\xm^{\top}\LM\xm
\end{equation*}
with a probability of at least $1-\delta$.
First, this approach fails to account for node attributes. Second, the computation of the ER of all edges in $\G$ is rather costly. Even the approximate algorithms \cite{spielman2008graph,10.1145/3580305.3599323} struggle to cope with medium-sized graphs. Besides, the edge sampling strategy relies on a large $n_r$ as it will repeatedly pick the same edges.

\stitle{ER Approximation on $\G_w$}
To this end, we first conduct a rigorous theoretical analysis in Lemma \ref{lem:ER} and disclose that the ER value of each edge $e_{i,j}$ in $\G_w$ is roughly proportional to $\frac{1}{d_w(v_i)} + \frac{1}{d_w(v_j)}$.
\begin{lemma}\label{lem:ER}
Let $\G_w=(\V,\EDG_w)$ be a weighted graph whose node degrees are defined as in Eq. \eqref{eq:d-weight}. The ER $r_w(e_{i,j})$ of each edge $e_{i,j}\in \EDG_w$ is bounded by
\begin{small}
\begin{equation*}
\frac{1}{2} \left( \frac{1}{d_w(v_i)}+\frac{1}{d_w(v_j)} \right) \le r_w(e_{i,j}) \le \frac{1}{1-\lambda_2} \left( \frac{1}{d_w(v_i)}+\frac{1}{d_w(v_j)} \right),
\end{equation*}
\end{small}
where $\lambda_2\le 1$ stands for the second largest eigenvalue of the normalized adjacency matrix of $\G_w$.
\begin{proof}
We defer the proof to Appendix~\ref{sec:proof-ER}.
\end{proof}
\end{lemma}
The above finding implies that we can leverage $\frac{1}{d_w(v_i)} + \frac{1}{d_w(v_j)}$ as an estimation of the ER of edge $e_{i,j}$ on $\G_w$, which roughly reflects the relative importance of edges.

\stitle{Edge Ranking and Sparsification of $\G_w$}
On this basis, in lieu of sampling edges in $\G_w$ for sparsified graph construction, we resort to ranking edges in ascending order as per their centrality values $C_w(e_{i,j})$ defined by
\begin{equation}\label{eq:e-centrality}
C_w(e_{i,j})=w(e_{i,j})\cdot \left(\frac{1}{d_w(v_i)} + \frac{1}{d_w(v_j)}\right),
\end{equation}
which intuitively quantifies the total importance of edge $e_{i,j}$ among all edges incident to $v_i$ and $v_j$. Afterwards, given a sparsification ratio $\rho$, we delete a subset $\EDG_{\textnormal{rm}}$ of edges with $m\rho$ lowest centrality values from $\G_w$ and construct the sparsified adjacency matrix $\AM^{\circ}$ as follows:
\begin{equation}
\forall{e_{i,j}\in \EDG_w\setminus \EDG_{\textnormal{rm}}}\quad \AM^{\circ}_{i,j} = w(e_{i,j}).
\end{equation}

The pseudo-code and complexity analysis are in Appendix~\ref{sec:sparsify-add}.








%% file: tex/experiments.tex
\begin{table}[!t]
\centering
\renewcommand{\arraystretch}{1.0}
\begin{footnotesize}
\caption{Statistics of Datasets ($\textnormal{K}=10^3\text{\ and\ }\textnormal{M}=10^6$).}\label{tbl:exp-data}
\vspace{-3ex}
\resizebox{\columnwidth}{!}{%
\begin{tabular}{l|r|r|r|c|c|c}
	\hline
	{\bf Dataset} & \multicolumn{1}{c|}{\bf $n$ } & \multicolumn{1}{c|}{\bf $m$ } & \multicolumn{1}{c|}{\bf $d$ } & \multicolumn{1}{c|}{\bf $|\Y|$} & \multicolumn{1}{c|}{\bf ${m}/{n}$} & \multicolumn{1}{c}{\bf $HR$} \\
	\hline
	\hline
    {\em Photo} \cite{shchur2018pitfalls} & 7.7K & 238.2K & 745 & 8 & 31.1 & 0.83 \\
    {\em WikiCS} \cite{mernyei2020wiki} & 11.7K &  431.7K & 300 & 10 & 36.9 & 0.65 \\
    {\em Reddit2} \cite{zeng2019graphsaint} & 233K & 23.2M & 602 & 41 & 99.6 & 0.78\\
    {\em Amazon2M} \cite{chiang2019cluster} & 2.45M & 61.9M & 100 & 47 & 25.3 & 0.81 \\
    \hline
    {\em Squirrel} \cite{pei2019geom} & 5.2K &  396.9K & 2.1K & 5 & 76.3 & 0.22 \\
    {\em Penn94} \cite{hu2020open} & 41.6K &   1.4M & 128 & 2 & 32.8 & 0.47 \\
    {\em Ogbn-Proteins} \cite{hu2020open} &  132.5K & 39.6M & 8 & 112  & 298.5 & 0.38 \\
    {\em Pokec}~\cite{lim2021large} &1.6M &   30.6M & 65 & 2 & 18.8 & 0.45 \\
    \hline
\end{tabular}%
}
\vspace{-2ex}
\end{footnotesize}
\end{table}

\section{Experiments}

\subsection{Experimental Setup}
\vspace{-1ex}
\stitle{Datasets}
Table~\ref{tbl:exp-data} lists the statistics of 8 benchmark HDGs ($m/n\ge 18$) tested in our experiments, which are of diverse types and varied sizes. $|\Y|$ symbolizes the distinct number of class labels of nodes in $\G$. The {\em homophily ratio} (HR) of $\G$ is defined as the fraction of homophilic edges linking same-class nodes \cite{zhu2020beyond}. We refer to a graph with $HR \ge 0.5$ as homophilic and as heterophilic if $HR < 0.5$. Particularly, datasets {\em Photo}~\cite{shchur2018pitfalls}, {\em WikiCS}~\cite{mernyei2020wiki}, {\em Reddit2}~\cite{zeng2019graphsaint}, and {\em Amazon2M}~\cite{chiang2019cluster} are homophilic graphs, whereas {\em Squirrel}~\cite{pei2019geom}, {\em Penn94}~\cite{hu2020open}, {\em Ogbn-Proteins}~\cite{hu2020open}, and {\em Pokec}~\cite{lim2021large} are heterophilic graphs. {\em Amazon2M} and {\em Pokec} are two large HDGs with millions of nodes and tens of millions of edges. More details of the datasets and train/validation/test splits can be found in Appendix~\ref{sec:datasets-more}.

\begin{table*}[!t]
\centering
\renewcommand{\arraystretch}{1.0}
\caption{Node classification results (\% test accuracy) of different GNN backbones with and without \algo on homophilic and heterophilic graphs. We conduct 10 trials and report mean accuracy and standard deviation over the trials.}\vspace{-3mm}
\begin{small}
\addtolength{\tabcolsep}{-0.25em}
\begin{tabular}{c|c|c|c|c|c|c|c|c}
\hline
\multirow{1}{*}{\bf Method} & \multicolumn{1}{c|}{\bf{ {\em Photo}}} & \multicolumn{1}{c|}{\bf{ {\em WikiCS}}}  & \multicolumn{1}{c|}{\bf{ {\em Reddit2}}}  & \multicolumn{1}{c|}{\bf{ {\em Amazon2M}}}  & \multicolumn{1}{c|}{\bf{ {\em Squirrel}}} & \multicolumn{1}{c|}{\bf{ {\em Penn94}}}  & \multicolumn{1}{c|}{\bf{ {\em Ogbn-Proteins}}}  & \multicolumn{1}{c}{\bf{ {\em Pokec}}} \\ \cline{1-9}
\hline
\hline
 GCN &94.63±0.15&84.05±0.76	&92.58±0.03&74.12±0.19&54.85±2.02&75.9±0.74&69.75±0.6 & {\bf 75.47±1.36}\\ 
 GCN + \algo &\bf 94.92±0.45&{\bf 84.62±0.53} &{\bf 94.86±0.22}&{\bf 76.14±0.23}&{\bf 73.48±1.61}& \bf 76.06±0.43 &{\bf 73.79±0.76}&75.01±0.27\\  \hline
 GAT &93.84±0.46&83.74±0.75&OOM&OOM&55.70±3.26&71.09±1.35&OOM& 73.20±7.02\\ 
 GAT + \algo &{\bf 94.58±0.12}&{\bf 84.97±0.84}&	{\bf95.97±0.04}&{\bf 59.16±0.36}&{\bf 72.99±2.81}&{\bf 71.19±0.78} &{\bf 74.94±0.25}&{\bf74.26±0.94}\\ \hline
 SGC &93.29±0.79&83.47±0.83&94.78±0.02&59.86±0.04&52.18±1.49&56.77±0.14&70.33±0.04& {\bf 67.40±5.56}\\ 
 SGC + \algo &{\bf 94.93±0.39}&{\bf 83.97±0.71}&{\bf95.65±0.02}&{\bf73.39±0.35}&{\bf 72.32±2.72}&{\bf 71.02±0.53}&{\bf74.31±0.42}&62.06±0.52\\  \hline
 APPNP &94.95±0.33&85.04±0.60&90.86±0.19&65.51±0.36&54.47±2.06&69.25±0.38&75.19±0.58&62.79±0.11\\ 
 APPNP + \algo &\bf 95.42±0.53&{\bf 85.19±0.56}&{\bf95.34±0.18}&{\bf69.81±0.24}&{\bf 73.24±1.38}&{\bf 71.08±0.62}& {\bf 75.52±0.32 }& {\bf 67.03±0.27}\\  \hline
 GCNII &95.12±0.12&85.13±0.56&94.66±0.07& OOM         &53.13±4.29&  74.97±0.35 &73.11±1.93&76.49±0.88\\ 
 GCNII + \algo & \bf 95.54±0.44&{\bf 85.42±0.60}&{\bf 96.62±0.08}&{\bf77.83±0.62}&{\bf72.89±2.45}& {\bf 75.84±3.13} &{\bf75.34±1.33}&{\bf77.64±0.32}\\  \hline
\end{tabular}
\end{small}
\label{tbl:node-class}
\vspace{0ex}
\end{table*}

\stitle{Baselines and Configurations} 
We adopt five popular MP-GNN architectures, GCN~\cite{kipf2016semi}, GAT~\cite{velivckovic2018graph}, SGC~\cite{wu2019simplifying}, APPNP~\cite{gasteiger2018predict}, and GCNII~\cite{chen2020simple} as the baselines and backbones to validate \algo in semi-supervised node classification tasks (Section \ref{sec:classification}). To demonstrate the superiority of \algo, we additionally compare \algo against other GDA techniques 
in Section~\ref{sec:GDA}
its variants with other feature expansion 
and graph sparsification strategies 
in Section~\ref{sec:ablation}. The implementation details and hyper-parameter settings can be found in Appendix ~\ref{sec:baseline-more}.
All experiments are conducted on a Linux machine with an NVIDIA Ampere A100 GPU (80GB RAM), AMD EPYC 7513 CPU (2.6 GHz), and 1TB RAM. Source codes can be accessed at \url{https://github.com/HKBU-LAGAS/TADA}. 

\subsection{ Semi-Supervised Node Classification}\label{sec:classification}
\vspace{-1ex}
\stitle{Effectiveness} In this set of experiments, we compare \algo-augmented GCN, GAT, SGC, APPNP, and GCNII models against their vanilla versions in terms of semi-supervised node classification. Table~\ref{tbl:node-class} reports their test accuracy results on 8 HDG datasets. OOM represents that the model fails to report results due to the out-of-memory issue. It can be observed that \algo consistently improves the baselines in accuracy on both homophilic and heterophilic graphs in almost all cases. Notably, on the {\em Squirrel} dataset, the five backbones are outperformed by their \algo counterparts with significant margins of $17.29\%$-$20.14\%$ in testing accuracy. The reason is that {\em Squirrel} is endowed with uninformative nodal attributes, and by contrast, its structural features are more conducive for node classification. By expanding original node features with high-quality structure embeddings (Module I in \algo), \algo is able to overcome such problems and advance the robustness and effectiveness of GNNs. In addition, on {\em Reddit2} and {\em Ogbn-Proteins} with average degrees ($m/n$) over hundreds, \algo also yields pronounced improvements in accuracy, i.e., $2.28\%$ and $2.94\%$ for GCN, as well as $1.96\%$ and $2.23\%$ for GCNII, respectively.
This demonstrates the effectiveness of our graph sparsification method (Module II in \algo) in reducing noisy edges and mitigating over-smoothing issues particularly in graphs ({\em Reddit2} and {\em Ogbn-Proteins}) consisting of a huge number of edges (analysed in Section~\ref{sec:HDG-limitations}). On the rest HDGs, almost all GNN backbones see accuracy gains with \algo. Two exceptions occur on heterophilic HDG {\em Pokec}, where GCN and SGC get high standard deviations ($1.36\%$ and $5.56\%$) in accuracy while GCN+\algo and SGC+\algo attenuate average accuracies but increase their performance stability.

\begin{figure}[!t]
\vspace{-2ex}
\centering
\begin{small}
\subfloat[{\em Reddit2}]{
\begin{tikzpicture}[scale=1]
\begin{axis}[
    height=\columnwidth/2.8,
    width=\columnwidth/1.7,
    ybar=0.1pt,
    bar width=0.2cm,
    enlarge x limits=true,
    ylabel={\em running time} (ms),
    symbolic x coords={2,4,6,8},
    xticklabels={GCN,SGC,APPNP,GCNII},
    xtick=data,
    ymin=1,
    ymax=150,
    ytick={1,10,100},
    yticklabels={$10^0$,$10^1$,$10^2$},
    ymode=log,
    y label style = {font=\footnotesize},
    xlabel near ticks,
    x label style = {font=\small},
    yticklabel style = {font=\tiny},
    log basis y={10},
    every axis y label/.style={font=\small,at={{(0.25,1.0)}},right=-1mm,above=0mm},
        label style={font=\scriptsize},
        tick label style={font=\scriptsize},
    ]
\addplot [black, fill=B3, postaction={ pattern=north east lines }] coordinates {(2,53.38) (4,25.92) (6,125.55) (8,77.57)};
\addplot [black, fill=O1, postaction={pattern=grid}] coordinates {(2,23.96) (4,18.08) (6,57.33)  (8,49.51)};
\end{axis}
\end{tikzpicture}\hspace{1mm}%
}%
\subfloat[{\em Ogbn-Proteins}]{
\begin{tikzpicture}[scale=1]
\begin{axis}[
    height=\columnwidth/2.8,
    width=\columnwidth/1.7,
    ybar=0.1pt,
    bar width=0.2cm,
    enlarge x limits=true,
    ylabel={\em running time} (ms),
    symbolic x coords={2,4,6,8},
    xticklabels={GCN,SGC,APPNP,GCNII},
    xtick=data,
    yticklabel style = {font=\tiny},
    ymin=1,
    ymax=300,
    ytick={1,10,100},
    ymode=log,
    log basis y={10},
    y label style = {font=\footnotesize},
    xlabel near ticks,
    x label style = {font=\small},
    every axis y label/.style={font=\small,at={{(0.25,1.0)}},right=-1mm,above=0mm},
        label style={font=\scriptsize},
        tick label style={font=\scriptsize},
    ]
\addplot [black, fill=B3, postaction={ pattern=north east lines }] coordinates {(2,210.59) (4,18.82) (6,219.20) (8,211.31)};
\addplot [black, fill=O1, postaction={pattern=grid}] coordinates {(2,52.26) (4,16.68) (6,106.46) (8,42.368)};
\end{axis}
\end{tikzpicture}\hspace{0mm}%
}%
\vspace{-4mm}
\end{small}
\caption{Training time per epoch} \label{fig:efficiency-traintime}
\vspace{-2ex}
\end{figure}
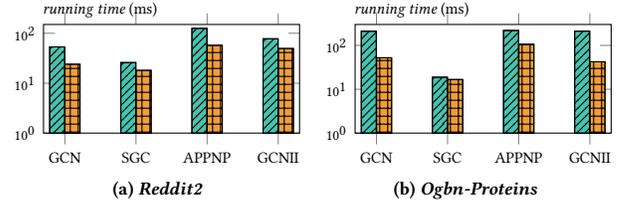

\begin{figure}[!t]
\centering
\begin{small}
\begin{tikzpicture}
    \begin{customlegend}[legend columns=2,
        legend entries={Baseline, Baseline+\algo},
        legend columns=-1,
        area legend,
        legend style={at={(0.55,1.15)},anchor=north,draw=none,font=\small,column sep=0.3cm}]
        \addlegendimage{black, fill=B3, postaction={ pattern=north east lines }}
        \addlegendimage{black, fill=O1, postaction={pattern=grid}}
    \end{customlegend}
\end{tikzpicture}
\\[-\lineskip]
\vspace{-4mm}
\subfloat[{\em Reddit2}]{
\begin{tikzpicture}[scale=1]
\begin{axis}[
    height=\columnwidth/2.8,
    width=\columnwidth/1.7,
    ybar=0.1pt,
    bar width=0.2cm,
    enlarge x limits=true,
    ylabel={\em running time} (ms),
    symbolic x coords={2,4,6,8},
    xticklabels={GCN,SGC,APPNP,GCNII},
    xtick=data,
    ymin=10,
    ymax=1,
    ymode=log,
    y label style = {font=\footnotesize},
    xlabel near ticks,
    x label style = {font=\small},
    yticklabel style = {font=\tiny},
    log origin y=infty,
    log basis y={10},
    every axis y label/.style={font=\small,at={{(0.25,1.0)}},right=-1mm,above=0mm},
        label style={font=\scriptsize},
        tick label style={font=\scriptsize},
    ]
\addplot [black, fill=B3, postaction={ pattern=north east lines }] coordinates {(2,0.31) (4,0.12) (6,0.33)  (8,0.66)};
\addplot [black, fill=O1, postaction={pattern=grid}] coordinates {(2,0.44) (4,0.24) (6,0.44)  (8,0.72)};
\end{axis}
\end{tikzpicture}\hspace{1mm}%
}%
\subfloat[{\em Ogbn-Proteins}]{
\begin{tikzpicture}[scale=1]
\begin{axis}[
    height=\columnwidth/2.8,
    width=\columnwidth/1.7,
    ybar=0.1pt,
    bar width=0.2cm,
    enlarge x limits=true,
    ylabel={\em running time} (ms),
    symbolic x coords={2,4,6,8},
    xticklabels={GCN,SGC,APPNP,GCNII},
    xtick=data,
    yticklabel style = {font=\tiny},
    ymin=0.1,
    ymax=150,
    ymode=log,
    log basis y={10},
    log origin y=infty,
    y label style = {font=\footnotesize},
    xlabel near ticks,
    x label style = {font=\small},
    every axis y label/.style={font=\small,at={{(0.25,1.0)}},right=-1mm,above=0mm},
        label style={font=\scriptsize},
        tick label style={font=\scriptsize},
    ]
\addplot [black, fill=B3, postaction={ pattern=north east lines }] coordinates {(2,94.91) (4,0.69) (6,95.16) (8,95.52)};
\addplot [black, fill=O1, postaction={pattern=grid}] coordinates {(2,0.78) (4,0.17) (6,0.48) (8,1.11)};
\end{axis}
\end{tikzpicture}\hspace{0mm}%
}%
\vspace{-4mm}
\end{small}
\caption{Inference time per epoch} \label{fig:efficiency-infertime}
\vspace{-2ex}
\end{figure}
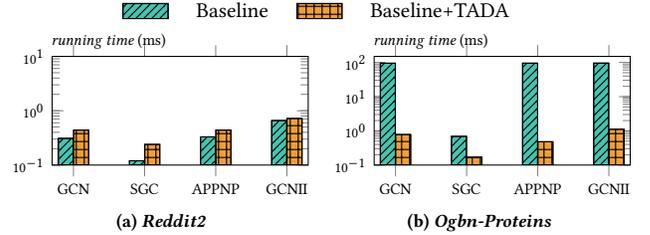

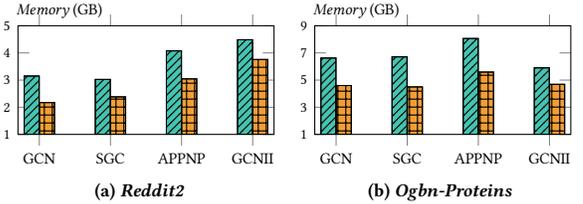
\begin{figure}[!t]
\vspace{-2ex}
\centering
\begin{small}
\subfloat[{\em Reddit2}]{
\begin{tikzpicture}[scale=1]
\begin{axis}[
    height=\columnwidth/2.8,
    width=\columnwidth/1.7,
    ybar=0.1pt,
    bar width=0.2cm,
    enlarge x limits=true,
    ylabel={\em Memory} (GB),
    symbolic x coords={2,4,6,8},
    xticklabels={GCN,SGC,APPNP,GCNII},
    xtick=data,
    ymin=1,
    ymax=5,
    ytick={1,2,3,4,5},
    y label style = {font=\footnotesize},
    xlabel near ticks,
    x label style = {font=\small},
    yticklabel style = {font=\tiny},
    every axis y label/.style={font=\small,at={{(0.25,1.0)}},right=-3mm,above=0mm},
        label style={font=\scriptsize},
        tick label style={font=\scriptsize},
    ]
\addplot [black, fill=B3, postaction={ pattern=north east lines }] coordinates {(2,3.1498) (4,3.0237) (6,4.0776) (8,4.4853) };
\addplot [black, fill=O1, postaction={pattern=grid}] coordinates {(2,2.1717) (4,2.3825) (6,3.0446)  (8,3.7577) };
\end{axis}
\end{tikzpicture}\hspace{1mm}%
}%
\subfloat[{\em Ogbn-Proteins}]{
\begin{tikzpicture}[scale=1]
\begin{axis}[
    height=\columnwidth/2.8,
    width=\columnwidth/1.7,
    ybar=0.1pt,
    bar width=0.2cm,
    enlarge x limits=true,
    ylabel={\em Memory} (GB),
    symbolic x coords={2,4,6,8},
    xticklabels={GCN,SGC, APPNP,GCNII},
    xtick=data,
    yticklabel style = {font=\tiny},
    ymin=1,
    ymax=9,
    ytick={1,3,5,7,9},
    y label style = {font=\footnotesize},
    xlabel near ticks,
    x label style = {font=\small},
    every axis y label/.style={font=\small,at={{(0.25,1.0)}},right=-3mm,above=0mm},
        label style={font=\scriptsize},
        tick label style={font=\scriptsize},
    ]
\addplot [black, fill=B3, postaction={ pattern=north east lines }] coordinates {(2,6.6249) (4,6.7123) (6,8.062) (8,5.902)};
\addplot [black, fill=O1, postaction={pattern=grid}] coordinates {(2,4.60) (4,4.4962) (6,5.59) (8,4.6872)};
\end{axis}
\end{tikzpicture}\hspace{2mm}%
}%
\vspace{-4mm}
\end{small}
\caption{Maximum GPU Memory Usage} \label{fig:memory}
\vspace{-2ex}
\end{figure}

\stitle{Efficiency}
To assess the effectiveness of \algo in the reduction of GNNs' feature aggregation overhead on HDGs, Figures~\ref{fig:efficiency-infertime}, ~\ref{fig:efficiency-traintime}, and ~\ref{fig:memory} plot the inference times and training times per epoch (in milliseconds), as well as the maximum memory footprints (in GBs) needed by four GNN backbones (GCN, SGC, APPNP, and GCNII) and their \algo counterparts on a heterophilic HDG {\em Ogbn-Proteins} and a homophilic HDG {\em Reddit2}. We exclude GAT as it incurs OOM errors on these two datasets, as shown in Table~\ref{tbl:node-class}. From Figure~\ref{fig:efficiency-infertime}, we note that 
on {\em Ogbn-Proteins}, \algo is able to speed up the inferences of GCN, APPNP, and GCNII to $121.7\times$, $198.2\times$, and $86\times$ faster, respectively, whereas on {\em Reddit2} \algo achieves comparable runtime performance to the vanilla GNN models.
This reveals that {\em Reddit2} and {\em Ogbn-Proteins} contains substantial noisy or redundant edges that can be removed without diluting the results of GNNs if \algo is included.
\eat{
The reason is that, compared to {\em Reddit2}, {\em Ogbn-Proteins} is a heterophilic HDG containing more noisy or redundant edges that can be removed without diluting results. 
}
Apart from the inference, \algo can also slightly expedite the training in the presence of Module I and Module II (see Figure~\ref{fig:efficiency-traintime}), indicating the high efficiency of our techniques developed in \algo.
In addition to the superiority in computational time, it can be observed from Figure~\ref{fig:memory} that \algo leads to at least a $24\%$ and $16\%$ reduction in memory consumption compared to the vanilla GNN models.

In a nutshell, \algo successfully addresses the technical challenges of GNNs on HDGs as remarked in Section~\ref{sec:HDG-limitations}.
Besides, we refer interested readers to Appendix~\ref{sec:exp-LDGs} for the empirical studies of \algo on low-degree graphs.






\begin{table}[!t]
\centering
\renewcommand{\arraystretch}{1.0}
\caption{Comparison with GDA Baselines.}\vspace{-3mm}
\begin{small}
\addtolength{\tabcolsep}{-0.25em}
\resizebox{\columnwidth}{!}{%
\begin{tabular}{c|cc| cc}
\hline
\multirow{2}{*}{\bf Method} & \multicolumn{2}{c|}{\bf{ {\em Reddit2}}} & \multicolumn{2}{c}{\bf{ {\em Ogbn-Proteins}}}  \\ \cline{2-5}
& Acc (\%) & Trng. / Inf. (ms)  & Acc (\%) & Trng. / Inf. (ms) \\ 
\hline
\hline
GCN    	&      $92.58_{\pm0.03}$    &	53.4 / 0.31 &$69.75_{\pm0.6}$& 	210.59 / 94.01 	\\
GCN+DropEdge    & \underline{$93.59_{\pm0.05}$}& \underline{49.51 / 0.31} & $61.46_{\pm 3.33}$ & \underline{62.65 / 93.53}\\
GCN+GraphMix    &	$92.60_{\pm 0.07}$	&128.58 / 0.38 & \underline{$72.41_{\pm 1.34}$} &441.23/93.95\\
GCN+\algo &     { $\bf 94.86_{\pm 0.22}$}    &		{\bf 24 / 0.44} &$\bf 73.79_{\pm0.76}$& 	{\bf 52.26 / 0.78}		\\ \hline
GCNII    & {$94.66_{\pm 0.07}$}&		125.6 / 0.66	& \underline{$73.11_{\pm1.93}$} & 	211.31 / 95.52	\\ 
GCNII+DropEdge   & \underline{$96.23_{\pm 0.05}$} & \underline{72.39 / 0.66} & $60.50_{\pm 5.42}$	& \underline{67.45 / 95.39}\\  
GCNII+GraphMix    &	$96.19_{\pm 0.05}$	&172.66/0.72 &$63.75_{\pm 1.72}$&456.59 / 95.34\\  
GCNII+\algo &    { $\bf96.62_{\pm 0.08}$}     &	 {\bf 49.5 / 0.72}	&$\bf 75.34_{\pm1.33}$& {\bf 42.68 / 1.11}	\\ 
\hline
\end{tabular}
}
{\textsuperscript{*}\footnotesize{Best is bolded and runner-up \underline{underlined}.}\hfill}
\end{small}
\label{tbl:data-aug-baselines}
\vspace{-3ex}
\end{table}

\subsection{Comparison with GDA Baselines}\label{sec:GDA}
This set of experiments evaluates the effectiveness \algo in improving GNNs' performance against other popular GDA techniques: DropEdge~\cite{rong2019dropedge} and GraphMix~\cite{verma2021graphmix}. Table~\ref{tbl:data-aug-baselines} presents the test accuracy results, training and inference times per epoch (in milliseconds) achieved by two GNN backbones GCN and GCNII and their augmented versions on {\em Ogbn-Proteins} and {\em Reddit2}. 
We can make the following observations. First, \algo+GCN and \algo+GCNII dominate all their competitors on the two datasets, respectively, in terms of classification accuracy as well as training and inference efficiency. On {\em Ogbn-Proteins}, we can see that the classification performance of GCN+DropEdge, GCNII+DropEdge, and GCNII+GraphMix is even inferior to the baselines, while {taking longer training and inference times.}, which is consistent with our analysis of the limitations of existing GDA methods on HDGs in Sections~\ref{sec:intro} and ~\ref{sec:related-work}.

\subsection{Ablation Study}\label{sec:ablation}

Table~\ref{tbl:ablation} presents the ablation study of \algo with GCN as the backbone model on 
{\em Reddit2} and {\em Ogbn-Proteins}.
More specifically, we conduct the ablation study in three dimensions. Firstly, we start with the vanilla GCN and incrementally apply components Count-Sketch, RWR-Sketch (Module I), and our graph sparsification technique (Module II) to the GCN. Notice that Module II is built on the output of Module I, and, thus, can only be applied after it. From Table~\ref{tbl:ablation}, we can observe that each component in \algo yields notable performance gains in node classification on the basis of the prior one, which exhibits the non-triviality of the modules to the effectiveness of \algo.

On the other hand, to demonstrate the superiority of our hybrid sketching approach introduced in Section~\ref{sec:sketched-embedding}, we substitute Count-Sketch and RWR-Sketch in Module I with random projection~\cite{li2006very}, $k$-SVD~\cite{halko2011finding}, DeepWalk~\cite{perozzi2014deepwalk}, node2vec~\cite{grover2016node2vec}, and LINE~\cite{tang2015line}, respectively, while fixing Module II. That is, we employ the random projections of adjacency matrix $\AM$, the top-$k$ singular vectors (as in \cite{sun2023feature}), or the node embeddings output by DeepWalk, node2vec, and LINE as $\AM^{\prime}$ for the generation of structure embeddings. As reported in Table~\ref{tbl:ablation}, all these five approaches obtain inferior classification results compared to \algo with Count-Sketch + RWR-Sketch on 
{\em Reddit2} and {\em Ogbn-Proteins}. 

Finally, we empirically study the effectiveness of our topology- and attribute-aware sparsification method in Section~\ref{sec:sparsification} (Module II) by replacing it with random sparsification(RS), $k$-Neighbor Spar~\cite{sadhanala2016graph}, SCAN~\cite{xu2007scan} and the DSpar~\cite{liu2023dspar}. Random sparsification removes edges randomly, and $k$-Neighbor Spar~\cite{sadhanala2016graph} samples at most $k$ edges for each neighbor. SCAN removes the edges with the lowest modified Jaccard similarity, while Dspar identifies the subset of dropped edges based on their estimated ER values in the original unweighted graph. {Table~\ref{tbl:ablation} shows that all these four variants are outperformed by \algo by a large margin.} On {\em Reddit2} and {\em Ogbn-Proteins}, \algo takes a lead of $0.89\%$ in classification accuracy compared to its best variant with $k$-Neighbor Spar.

\begin{table}[!t]
\centering
\renewcommand{\arraystretch}{1.1}
\caption{Ablation Study}\vspace{-3mm}
\begin{small}
\addtolength{\tabcolsep}{-0.25em}
{
\begin{tabular}{c|c|c}
\hline
\multirow{1}{*}{\bf Method}  & \multicolumn{1}{c}{\bf{ {\em Reddit2}}} & \multicolumn{1}{|c}{\bf{ {\em Obgn-Proteins}}} \\ \cline{2-2}
\hline
\hline
 GCN  & $92.58_{\pm 0.03}$ & $69.75_{\pm0.60}$\\ 
+ Count-Sketch  &  $93.81_{\pm0.88}$ & $72.90_{\pm2.14}$ \\ 
+ RWR-Sketch & 		$94.25_{\pm0.66}$ & $70.33_{\pm2.54}$\\ 
+ Module II (i.e., GCN+\algo)  &   $\bf 94.86_{\pm 0.22}$ & $\bf 73.79_{\pm0.76}$ \\ 
\hline
Random Projection (Module I)  		& {$93.99_{\pm0.74}$}	& $72.26_{\pm0.77}$	\\ 
$k$-SVD (Module I) & 		$93.36_{\pm0.51}$	&  $69.79_{\pm1.37}$	\\ 
DeepWalk (Module I) &  $94.48_{\pm0.34}$ & $72.56_{\pm0.94}$ \\
node2vec (Module I) & $94.47_{\pm0.41}$ &  \underline{$73.05_{\pm1.50}$} \\
LINE (Module I) &  \underline{$94.49_{\pm0.34}$}  & $72.49_{\pm1.66
}$ \\
\hline
RS (Module II) 	 &  	$91.04_{\pm0.04}$	& $72.54_{\pm1.11}$ \\ 
$k$-Neighbor Spar (Module II) & $93.97_{\pm0.7}$ & $72.90_{\pm1.47}$\\
SCAN (Module II) &$89.93_{\pm0.78}$ & $71.85_{\pm1.25}$ \\
DSpar (Module II) & 	 $93.58_{\pm0.08}$		& $72.75_{\pm1.11}$ \\ 
\hline
\end{tabular}
}\\
{\hspace{6mm}\textsuperscript{*}\footnotesize{Best is bolded and runner-up \underline{underlined}.}\hfill}
\end{small}
\label{tbl:ablation}
\vspace{-1ex}
\end{table}





 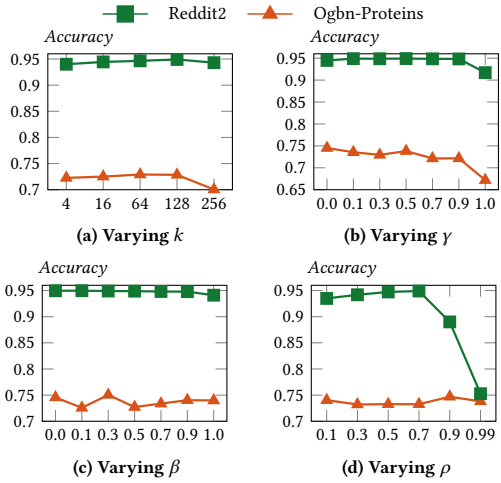
\begin{figure}[!t]
\centering
\begin{small}
\begin{tikzpicture}
    \begin{customlegend}[legend columns=2,
        legend entries={Reddit2,Ogbn-Proteins},
        legend style={at={(0.45,1.35)},anchor=north,draw=none,font=\footnotesize,column sep=0.2cm}]
    \addlegendimage{line width=0.2mm,mark size=3pt,mark=square*, color=my_green}
    \addlegendimage{line width=0.2mm,mark size=3pt,mark=triangle*, color=my_orange}
    
    \end{customlegend}
\end{tikzpicture}
\\[-\lineskip]
\vspace{-4ex}
\subfloat[Varying $k$]{
\begin{tikzpicture}[scale=1,every mark/.append style={mark size=2pt}]
    \begin{axis}[
        height=\columnwidth/2.5,
        width=\columnwidth/2.1,
        ylabel={\it Accuracy},
        xmin=0.5, xmax=5.5,
        ymin=0.7, ymax=0.96,
        xtick={1,2,3,4,5},
        ytick={0.7,0.75,0.8,0.85,0.9,0.95},
        xticklabel style = {font=\footnotesize},
        yticklabel style = {font=\footnotesize},
        xticklabels={4,16,64,128,256},
        yticklabels={0.7,0.75,0.8,0.85,0.9,0.95},
        every axis y label/.style={font=\footnotesize,at={(current axis.north west)},right=4mm,above=0mm},
        legend style={fill=none,font=\small,at={(0.02,0.99)},anchor=north west,draw=none},
    ]
    \addplot[line width=0.3mm, mark=triangle*, color=my_orange] 
        plot coordinates {
(1,	0.7226	)
(2,	0.7251	)
(3,	0.7291	)
(4,	0.7285	)
(5,	0.7004	)

    };

    \addplot[line width=0.3mm, mark=square*, color=my_green]  
        plot coordinates {
(1,0.9400)
(2,0.9444)
(3,	0.9465	)
(4,	0.9491	)
(5,	0.9430	)

    };
    
    \end{axis}
\end{tikzpicture}\hspace{4mm}\label{fig:vary-k}%
}%
\subfloat[Varying $\gamma$]{
\begin{tikzpicture}[scale=1,every mark/.append style={mark size=2pt}]
    \begin{axis}[
        height=\columnwidth/2.5,
        width=\columnwidth/2.1,
        ylabel={\it Accuracy},
        xmin=0.5, xmax=7.5,
        ymin=0.65, ymax=0.96,
        xtick={1,2,3,4,5,6,7},
        ytick={0.65,0.7,0.75,0.8,0.85,0.9,0.95},
        xticklabel style = {font=\footnotesize},
        yticklabel style = {font=\footnotesize},
        xticklabels={0.0,0.1,0.3,0.5,0.7,0.9,1.0},
        yticklabels={0.65,0.7,0.75,0.8,0.85,0.9,0.95},
        every axis y label/.style={font=\footnotesize,at={(current axis.north west)},right=4mm,above=0mm},
        legend style={fill=none,font=\small,at={(0.02,0.99)},anchor=north west,draw=none},
    ]
    \addplot[line width=0.3mm, mark=triangle*, color=my_orange] 
        plot coordinates {
(1,	0.7451	)       
(2,	0.7355	)
(3,	0.7295	)
(4,	0.7381	)
(5,	0.7214	)
(6,	0.7215	)
(7,	0.6719	)
    };

    \addplot[line width=0.3mm, mark=square*, color=my_green]  
        plot coordinates {
(1,	0.9448	)
(2,	0.9489	)
(3,	0.9487	)
(4,	0.9491	)
(5,	0.9485	)
(6,	0.9483	)
(7,	0.9170	)
    };
    
    \end{axis}
\end{tikzpicture}\hspace{4mm}\label{fig:vary-gamma}%
}

\vspace{-3mm}
\subfloat[Varying $\beta$]{
\begin{tikzpicture}[scale=1,every mark/.append style={mark size=2pt}]
    \begin{axis}[
        height=\columnwidth/2.5,
        width=\columnwidth/2.1,
        ylabel={\it Accuracy},
        xmin=0.5, xmax=7.5,
        ymin=0.7, ymax=0.96,
        xtick={1,2,3,4,5,6,7},
        ytick={0.7,0.75,0.8,0.85,0.9,0.95},
        xticklabel style = {font=\footnotesize},
        yticklabel style = {font=\footnotesize},
        xticklabels={0.0,0.1,0.3,0.5,0.7,0.9,1.0},
        yticklabels={0.7,0.75,0.8,0.85,0.9,0.95,1.0},
        every axis y label/.style={font=\footnotesize,at={(current axis.north west)},right=4mm,above=0mm},
        legend style={fill=none,font=\small,at={(0.02,0.99)},anchor=north west,draw=none},
    ]
    \addplot[line width=0.3mm, mark=triangle*, color=my_orange] 
        plot coordinates {
(1,	0.7456	)
(2,	0.7257	)
(3,	0.7504	)
(4,	0.7272	)
(5,	0.7337	)
(6,	0.7403	)
(7,	0.7398	)
    };

    \addplot[line width=0.3mm, mark=square*, color=my_green]  
        plot coordinates {
(1,0.9496  )
(2,	0.9497	)
(3,	0.9491	)
(4,	0.9488	)
(5,	0.9482	)
(6,	0.9478	)
(7, 0.9411  )
    };

    \end{axis}
\end{tikzpicture}\hspace{4mm}\label{fig:vary-beta}%
}%
\subfloat[Varying $\rho$]{
\begin{tikzpicture}[scale=1,every mark/.append style={mark size=2pt}]
    \begin{axis}[
        height=\columnwidth/2.5,
        width=\columnwidth/2.1,
        ylabel={\it Accuracy},
        xmin=0.5, xmax=6.5,
        ymin=0.7, ymax=0.96,
        xtick={1,2,3,4,5,6},
        ytick={0.7,0.75,0.8,0.85,0.9,0.95},
        xticklabel style = {font=\footnotesize},
        yticklabel style = {font=\footnotesize},
        xticklabels={0.1,0.3,0.5,0.7,0.9, 0.99},
        yticklabels={0.7,0.75,0.8,0.85,0.9,0.95,1.0},
        every axis y label/.style={font=\footnotesize,at={(current axis.north west)},right=4mm,above=0mm},
        legend style={fill=none,font=\small,at={(0.02,0.99)},anchor=north west,draw=none},
    ]
    \addplot[line width=0.3mm, mark=triangle*, color=my_orange] 
        plot coordinates {
(1,	0.7403	)
(2,	0.7321	)
(3,	0.7327	)
(4,	0.7326	)
(5,	0.7469	)
(6, 0.7381)
    };

    \addplot[line width=0.3mm, mark=square*, color=my_green]  
        plot coordinates {
(1,	0.9349	)
(2,	0.9419	)
(3,	0.9473	)
(4,	0.9491	)
(5,	0.89	)
(6, 0.7527 )
    };
    \end{axis}
\end{tikzpicture}\hspace{4mm}\label{fig:vary-rho}%
}%
\end{small}
 \vspace{-3mm}
\caption{Hyper-parameter Analysis.} \label{fig:parameter}
\vspace{-2ex}
\end{figure}

\subsection{Hyper-parameter Analysis}
This section empirically studies the sensitivity of TADA to hyper-parameters including the weight for structure embeddings $\gamma$ (Eq.~\eqref{eq:init-feat}), structure embedding dimension $k$ (Section~\ref{sec:sketched-embedding}), RWR-Sketch weight $\beta$ (Eq.~\eqref{eq:A-prime}), and sparsification ratio $\rho$ (Section~\ref{sec:sparsification}), on two datasets {\em Ogbn-Proteins} and {\em Reddit2}.

Figure~\ref{fig:vary-k} depict the node classification accuracy results of GCN+\algo when varying $k$ in $\{4, 16, 64, 128, 256\}$. We can make analogous observations on {\em Reddit2} and {\em Ogbn-Proteins}. That is, the performance of GCN+\algo first improves as $k$ is increased from $4$ to $128$ (more structure features are captured) and then undergoes a decline when $k=256$, as a consequence of over-fitting.

In Figure~\ref{fig:vary-gamma}, we plot the node classification accuracy values attained by GCN+\algo when $\gamma$ is varied from $0$ to $1.0$. Note that when $\gamma=0$ (resp. $\gamma=1.0$), the initial node features $\HM^{(0)}$ defined in Eq.~\eqref{eq:init-feat} will not embody structure features $\HM_{\textnormal{topo}}$ (resp. node attributes $\HM_{\textnormal{attr}}$). It can be observed that GCN+\algo obtains improved classification results on {\em Reddit2} when varying $\gamma$ from $0$ to $0.9$, whereas its performance on {\em Ogbn-Proteins} constantly downgrades as $\gamma$ enlarges. The degradation is caused by its heterophilic property and using its topological features for graph sprasification (Section~\ref{sec:sparsification}) will accidentally remove critical connections.  

From Figure~\ref{fig:vary-beta}, we can see that the best performance is achieved when $\beta=0.1$ and $\beta=0.3$ on {\em Reddit2} and {\em Ogbn-Proteins}, respectively, which validates the superiority of our hybrid sketching approach in Section~\ref{sec:sketched-embedding} over Count-Sketch or RWR-Sketch solely.

As displayed in Figure~\ref{fig:vary-rho}, on {\em Reddit2}, we can observe that GCN+\algo experiences an uptick in classification accuracy when excluding $10\%$-$70\%$ edges from $\G$ using Module II in \algo, followed by a sharp downturn when $\rho>70\%$. 
On {\em Ogbn-Proteins}, the best result is attained when $\rho=0.9$, i.e., $90\%$ edges are removed from $\G$.
The results showcase that Module II can accurately identify up to 70\%-90\% edges from $\G$ that are noisy or redundant and obtain performance enhancements.

\begin{figure}[t!]
\vspace{-3ex}
\centering
\subfloat[\small GCN]{
\includegraphics[width=4cm]{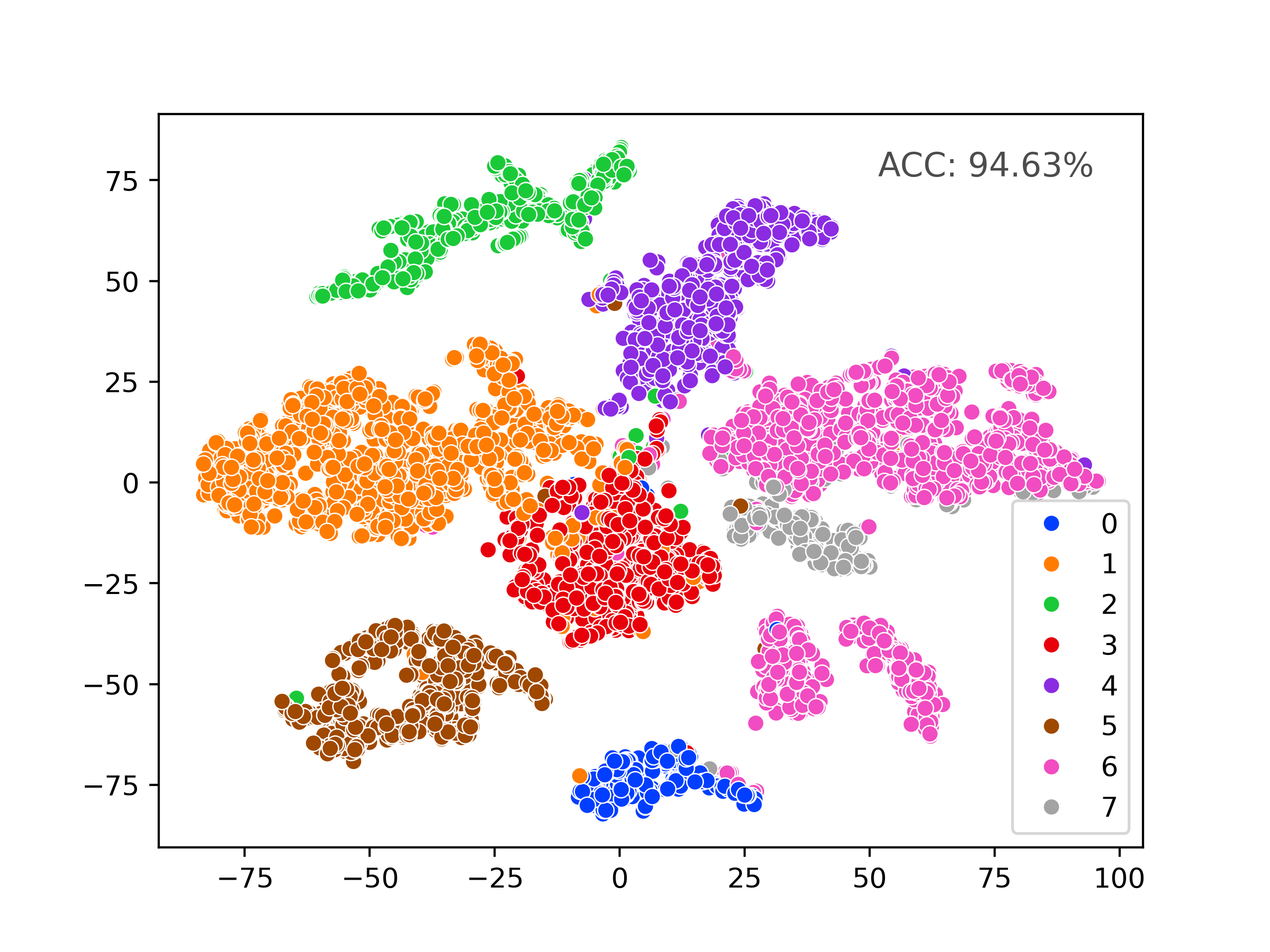}
\label{fig:gcn-vis}
}
\subfloat[\small GCN+\algo]{
\includegraphics[width=4cm]{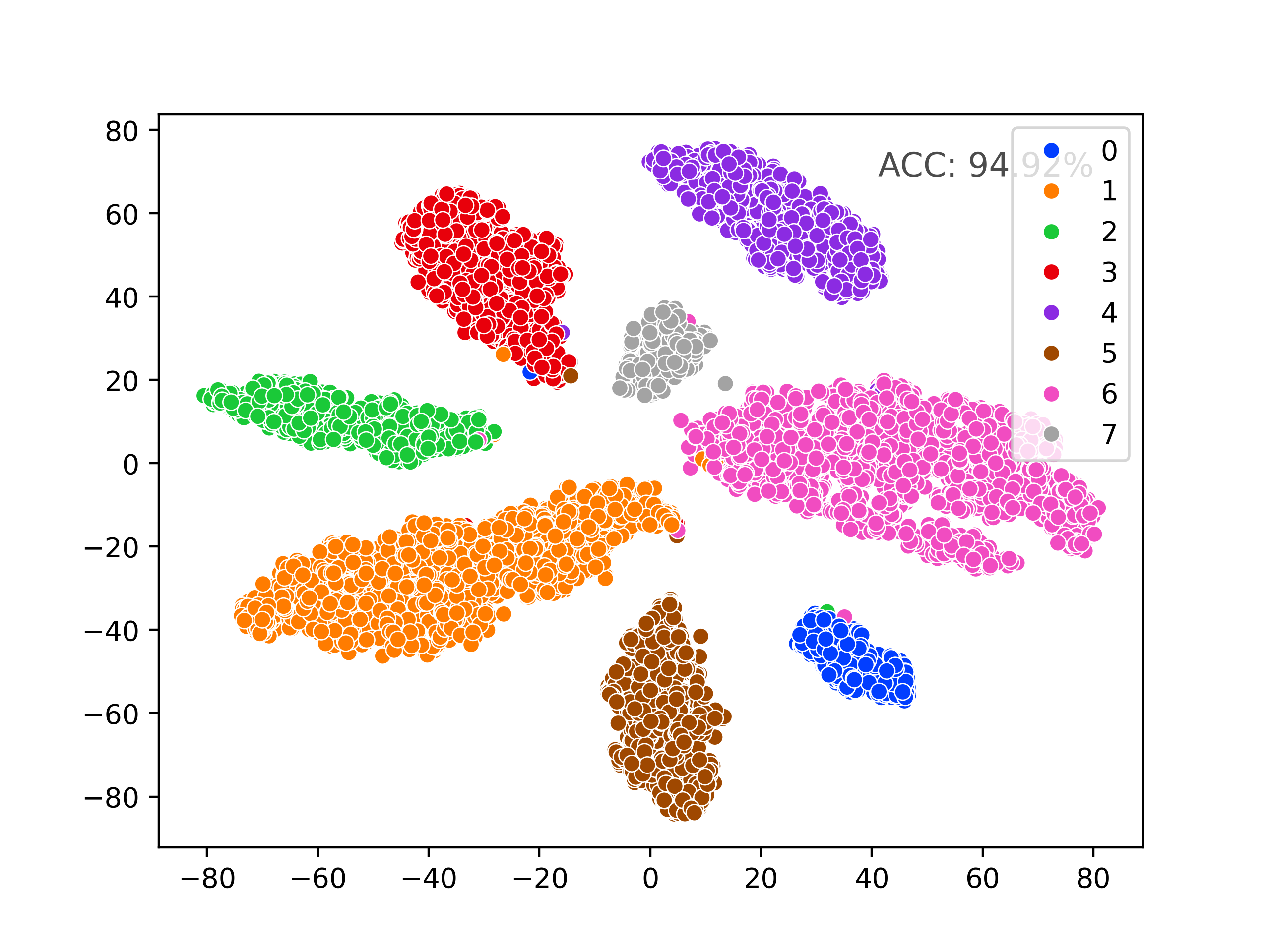}
\label{fig:gcn-tada-vis}
}\\
\vspace{-3mm}
\caption{The final node representations of {\em Photo} obtained by GCN and GCN+\algo. Nodes are colored by their labels.}\label{fig:vis}
\vspace{-2ex}
\end{figure}

\subsection{Visualization of \algo}
Figure~\ref{fig:vis} visualizes (using t-SNE~\cite{van2008visualizing}) the node representations of the {\em Photo} dataset at the final layers of GCN and GCN+\algo. Nodes with the same ground-truth labels will be in the same colors. In Figure~\ref{fig:gcn-tada-vis}, we can easily identify 8 classes of nodes  as nodes with disparate colors (i.e., labels) are all far apart from each other. In comparison, in Figure~\ref{fig:gcn-vis}, three groups of nodes with different colors are adjacent to each other with partial overlapping and some nodes even are positioned in other groups and distant from their true classes. These observations demonstrate that \algo can enhance the quality of nodes representations learned by GCN, and thus, yield the higher classification accuracy, as reported in Table~\ref{tbl:node-class}.

%% file: tex/proof.tex
\section{Proofs}\label{sec:proof}

{
\subsection{Proof of Lemma \ref{lem:ER}}\label{sec:proof-ER}
\begin{proof}
For ease of exposition, we re-define the notations here.
Let $\AM\in \mathbb{R}^{n\times n}$ be the weighted adjacency matrix of $\G$ wherein $\AM_{i,j}=w(v_i,v_j)$. Let $\DM\in \mathbb{R}^{n\times n}$ be a diagonal matrix where $\DM_{i,i}=\sum_{v_j\in \N(v_i)}{w(v_i,v_j)}$. The unnormalized graph Laplacian matrix is defined as $\LM=\DM-\AM$. Define matrices $\QM=\DM^{-\frac{1}{2}}\AM\DM^{-\frac{1}{2}}$ and $\PM=\DM^{-1}\AM$. We have $\QM=\DM^{\frac{1}{2}}\PM\DM^{-\frac{1}{2}}$ and $\LM = \DM^{\frac{1}{2}}\cdot\left(\IM-\QM\right)\cdot\DM^{\frac{1}{2}}$.

Let the eigendecomposition of $\QM$ be $\UM\boldsymbol{\Lambda}\UM^{\top}$. By definition,
\begin{align*}
\LM^{+} & = \DM^{-\frac{1}{2}}\cdot\left(\IM-\QM\right)^{+}\cdot\DM^{-\frac{1}{2}} = \DM^{-\frac{1}{2}}\cdot\left(\UM\frac{1}{\IM-\boldsymbol{\Lambda}}\UM^{\top}\right)\cdot\DM^{-\frac{1}{2}}.
\end{align*}
Since each eigenvalue $|\boldsymbol{\Lambda}_{i,i}|\le 1$, we have $\sum_{t=0}^{\infty}{\boldsymbol{\Lambda}_{i,i}^t}=\frac{1}{1-\boldsymbol{\Lambda}_{i,i}}$. Also, by the property of eigenvectors $\UM^{\top}\UM=\IM$, it is easy to derive that $\QM^t=\UM\boldsymbol{\Lambda}\UM^{\top}\UM\boldsymbol{\Lambda}\UM^{\top}\cdots\UM\boldsymbol{\Lambda}\UM^{\top}=\UM\boldsymbol{\Lambda}^t\UM^{\top}$. So,
\begin{align}
\LM^{+} & = \DM^{-\frac{1}{2}}\left(\UM \sum_{t=0}^{\infty}{\boldsymbol{\Lambda}^t}\UM^{\top}\right)\DM^{-\frac{1}{2}} = \DM^{-\frac{1}{2}}\sum_{t=0}^{\infty}{\QM^t}\DM^{-\frac{1}{2}} = \sum_{t=0}^{\infty}{\PM^t}\DM^{-1}\label{eq:pd}.
\end{align}
By the fact of $\QM^t=\UM\boldsymbol{\Lambda}^t\UM^{\top}$, for each $t$, we have
\begin{equation}
\PM^t\DM^{-1}=\DM^{-\frac{1}{2}}\cdot{\QM^t}\cdot\DM^{-\frac{1}{2}}=\DM^{-\frac{1}{2}}\cdot{\UM\boldsymbol{\Lambda}^t\UM^{\top}}\cdot\DM^{-\frac{1}{2}},
\end{equation}
which leads to
\begin{align*}
(\PM^t\DM^{-1})_{i,j}=\frac{\PM^t_{i,j}}{\DM_{j,j}}&=\sum_{k=1}^{n}{\frac{\UM_{i,k}}{\sqrt{\DM_{i,i}}}\cdot\frac{\UM_{j,k}}{\sqrt{\DM_{j,j}}}\cdot \boldsymbol{\Lambda}_{k,k}^t}.
\end{align*}
According to Lemma 12.2 in \cite{levin2017markov} and Lemma 3.1 in \cite{10.1145/3580305.3599323}, $\sqrt{2m}\DM^{-1}\UM_{:,1}=\mathbf{1}$. Thus, $\frac{\UM{i,k}}{\sqrt{\DM_{i,i}}}-\frac{\UM{j,k}}{\sqrt{\DM_{j,j}}}=\frac{1-1}{\sqrt{2m}}=0$, which leads to $(\PM^t\DM^{-1})_{i,j}=\sum_{k=2}^{n}{\frac{\UM_{i,k}}{\sqrt{\DM_{i,i}}}\cdot\frac{\UM_{j,k}}{\sqrt{\DM_{j,j}}}\cdot \boldsymbol{\Lambda}_{k,k}^t}$.

Further,
\begin{align*}
&(\PM^t\DM^{-1})_{i,i}+(\PM^t\DM^{-1})_{j,j}-2(\PM^t\DM^{-1})_{i,j}\\
&=\sum_{k=2}^{n}{\frac{\UM_{i,k}^2}{{\DM_{i,i}}}\cdot \boldsymbol{\Lambda}_{k,k}^t} + \sum_{k=2}^{n}{\frac{\UM_{j,k}^2}{{\DM_{j,j}}}\cdot \boldsymbol{\Lambda}_{k,k}^t} -2\sum_{k=2}^{n}{\frac{\UM_{i,k}}{\sqrt{\DM_{i,i}}}\cdot\frac{\UM_{j,k}}{\sqrt{\DM_{j,j}}}\cdot \boldsymbol{\Lambda}_{k,k}^t}\\
&=\sum_{k=2}^{n}{\left( \frac{\UM_{i,k}}{\sqrt{\DM_{i,i}}} - \frac{\UM_{j,k}}{\sqrt{\DM_{j,j}}}\right)^2\cdot \boldsymbol{\Lambda}_{k,k}^t}.
\end{align*}
In particular, when $t=0$, we have
\begin{equation}\label{eq:DD}
\frac{1}{\DM_{i,i}}+\frac{1}{\DM_{j,j}}=\sum_{k=2}^{n}{\left( \frac{\UM_{i,k}}{\sqrt{\DM_{i,i}}} - \frac{\UM_{j,k}}{\sqrt{\DM_{j,j}}}\right)^2}.
\end{equation}

In addition, by Eq. \eqref{eq:pd}, we have
\begin{align*}
&\LM^{+}_{i,i}+\LM^{+}_{j,j}-2\LM^{+}_{i,j} \\
&= \left(\sum_{t=0}^{\infty}{\PM^t}\DM^{-1}\right)_{i,i}+\left(\sum_{t=0}^{\infty}{\PM^t}\DM^{-1}\right)_{j,j}-2\left(\sum_{t=0}^{\infty}{\PM^t}\DM^{-1}\right)_{i,j}\\
&= \sum_{t=0}^{\infty}{\sum_{k=2}^{n}{\left( \frac{\UM_{i,k}}{\sqrt{\DM_{i,i}}} - \frac{\UM_{j,k}}{\sqrt{\DM_{j,j}}}\right)^2\cdot \boldsymbol{\Lambda}_{k,k}^t}} \\
&= \sum_{k=2}^{n}{\left( \frac{\UM_{i,k}}{\sqrt{\DM_{i,i}}} - \frac{\UM_{j,k}}{\sqrt{\DM_{j,j}}}\right)^2\cdot \sum_{t=0}^{\infty}{\boldsymbol{\Lambda}_{k,k}^t}},
\end{align*}
which leads to $\LM^{+}_{i,i}+\LM^{+}_{j,j}-2\LM^{+}_{i,j}=\sum_{k=2}^{n}{\left( \frac{\UM_{i,k}}{\sqrt{\DM_{i,i}}} - \frac{\UM_{j,k}}{\sqrt{\DM_{j,j}}}\right)^2\cdot \frac{1}{1-\boldsymbol{\Lambda}_{k,k}}}$.
Let $\lambda_2$ be the second largest eigenvalue of $\QM$. 
Since each eigenvalue's absolute value is not greater than $1$, by Eq. \eqref{eq:DD}, we have
\begin{align*}
& \LM^{+}_{i,i}+\LM^{+}_{j,j}-2\LM^{+}_{i,j} =\sum_{k=2}^{n}{\left( \frac{\UM_{i,k}}{\sqrt{\DM_{i,i}}} - \frac{\UM_{j,k}}{\sqrt{\DM_{j,j}}}\right)^2\cdot \frac{1}{1-\boldsymbol{\Lambda}_{k,k}}}\\
&\le \sum_{k=2}^{n}{\left( \frac{\UM_{i,k}}{\sqrt{\DM_{i,i}}} - \frac{\UM_{j,k}}{\sqrt{\DM_{j,j}}}\right)^2\cdot \frac{1}{1-\lambda_2}} =\frac{1}{1-\lambda_2}\cdot \left( \frac{1}{\DM_{i,i}}+\frac{1}{\DM_{j,j}} \right).
\end{align*}
And
\begin{align*}
\LM^{+}_{i,i}+\LM^{+}_{j,j}-2\LM^{+}_{i,j} &=\sum_{k=2}^{n}{\left( \frac{\UM_{i,k}}{\sqrt{\DM_{i,i}}} - \frac{\UM_{j,k}}{\sqrt{\DM_{j,j}}}\right)^2\cdot \frac{1}{1-\boldsymbol{\Lambda}_{k,k}}}\\
&\ge \sum_{k=2}^{n}{\left( \frac{\UM_{i,k}}{\sqrt{\DM_{i,i}}} - \frac{\UM_{j,k}}{\sqrt{\DM_{j,j}}}\right)^2\cdot \frac{1}{2}} =\frac{1}{2}\cdot \left( \frac{1}{\DM_{i,i}}+\frac{1}{\DM_{j,j}} \right).
\end{align*}
Therefore, the ER $r(e_{i,j})$ of edge $(v_i,v_j)$ satisifes
\begin{equation*}
\frac{1}{2}\cdot \left( \frac{1}{\DM_{i,i}}+\frac{1}{\DM_{j,j}} \right) \le r(e_{i,j}) \le \frac{1}{1-\lambda_2}\cdot \left( \frac{1}{\DM_{i,i}}+\frac{1}{\DM_{j,j}} \right)
\end{equation*}
The lemma is therefore proved.
\end{proof}
}

{
\subsection{Proof of Properties 1-3 in Section~\ref{sec:sketched-embedding}}\label{sec:proof-of-add-properties}
\begin{proof}
For Property 1, we let $\WM$ in Theorem \ref{lem:ARRW} be $\AM^{\top}$. Then, we have
\begin{align*}
&\mathbb{P}[ | \AM^{\prime}_{i}\cdot {\AM^{\prime}_j}^{\top} - \AM_i\AM_j^{\top} | < \epsilon ]\\
&=\mathbb{P}[ | \AM^{\prime}_{i}\cdot {\AM^{\prime}_j}^{\top} - |\N(v_i)\cap \N(v_j)| | < \epsilon ] > 1 - \delta.
\end{align*}
When $j=i$, we have
\begin{align*}
\mathbb{P}[ | \AM^{\prime}_{i}\cdot {\AM^{\prime}_i}^{\top} - \AM_i\AM_i^{\top} | < \epsilon ] & = \mathbb{P}[ | \|\AM^{\prime}_{i}\|^2_2 - \|\AM_i\|_2^2 | < \epsilon ] \\
& = \mathbb{P}[ | \|\AM^{\prime}_{i}\|^2_2 - d(v_i) | < \epsilon ] > 1-\delta.
\end{align*}

Since $\AM$ is symmetric, we have $\AM=\AM^{\top}$. Thus, $\WM$ in Theorem \ref{lem:ARRW} be $\AM^{\top}$
\begin{align*}
\mathbb{P}[ | \AM^{\prime}_i\cdot {\AM^{\prime}_j}^{\top} - \AM_i\cdot {\AM_j}^{\top} | < \epsilon ] & = \mathbb{P}[ | \AM^{\prime}_i\cdot {\AM^{\prime}_j}^{\top} - \AM^2_{i,j} | < \epsilon ] > 1 - \delta,
\end{align*}
which indicates that $\AM^{\prime}(\AM^{\prime})^{\top}$ is an approximation of $\AM^2$. As such, $(\AM^{\prime}(\AM^{\prime})^{\top})^t$ is naturally an approximation of $\AM^{2t}$. Property 2 is therefore proved.

For Property 3, by the definition of $\widehat{\AM^{\prime}}$, we first have
\begin{align*}
\widehat{\AM^{\prime}}_i = \frac{\AM^{\prime}_i}{\|\AM^{\prime}_i\|_2^2}\ \text{and}\ \widehat{\AM^{\prime}}_j = \frac{\AM^{\prime}_j}{\|\AM^{\prime}_j\|_2^2}.
\end{align*}
Recall that in Property 1, $\|\AM^{\prime}_{i}\|^2_2$ approximates $d(v_i)$, and in Property 2, $\AM^{\prime}_i(\AM^{\prime}_j)^{\top}=\AM^2_{i,j}=(\AM\AM^{\top})_{i,j}$. Hence,
\begin{align*}
\widehat{\AM^{\prime}}_i\cdot (\widehat{\AM^{\prime}}_j)^{\top} = \frac{\AM^{\prime}_i}{\|\AM^{\prime}_i\|_2^2}\cdot \left(\frac{\AM^{\prime}_j}{\|\AM^{\prime}_j\|_2^2}\right)^{\top}&\approx \DM^{-1}_{i,i}\cdot \AM^{\prime}_i(\AM^{\prime}_j)^{\top}\cdot \DM^{-1}_{j,j}\\
& = \DM^{-1}_{i,i}\cdot (\AM\AM^{\top})_{i,j}\cdot \DM^{-1}_{j,j} \\
& = (\DM^{-1}\AM)_{i}\cdot (\DM^{-1}\AM)_{j}^{\top}\\
& = \PM_i\cdot \PM_j^{\top},
\end{align*}
which further leads to Property 3.
\end{proof}
}

\eat{
\subsection{Proof of Theorem \ref{lem:1-WL}}
\begin{proof}
We prove the theorem using an example in Figure \ref{fig:WLtest}, which presents a 3-regular graph $\G$ with 8 nodes $v_1$-$v_8$. As shown in the figure, there are three isomorphism classes of nodes in $G$ with three different colors, i.e., $\V_i=\{v_1,v_2\}$, $\V_2=\{v_3,v_4,v_5,v_6\}$, and $\V_3=\{v_7,v_8\}$. As shown in previous works \cite{xu2018powerful,morris2019weisfeiler}, standard MP-GNNs cannot distinguish any pair of nodes in a regular graph since the computation trees rooted at all nodes in $\G$ are identical. 

In a message passing (feature aggregation) step of an MP-GNN that with $\AM$ as node features, nodes in $\V_1, \V_2$, and $\V_3$ will aggregate feature multisets $\{ \{\AM_1,\AM_2\}, \{\AM_4,\AM_6\}, \{\AM_3,\AM_5\} \}$, $\{\{\AM_1,\AM_2\}, \{\AM_4,\AM_6,\AM_3,\AM_5\} \{\AM_7,\AM_8\}\}$, and $\{\{\AM_7,\AM_8\}, \{\}\}$
\renchi{To-Do}.
\end{proof}
}

%% file: tex/other.tex
\section{Module I: Feature Expansion}\label{sec:sketch-add}

\subsection{Theoretical Analysis of Expanding Node Features with Adjacency Matrices}\label{sec:A-Feat}

Let $\boldsymbol{\Phi}$ be the feature space in standard GNNs. As per Eq. \eqref{eq:poly}, $\boldsymbol{\Phi}$ can be formulated by the multiplication of graph structure matrices (e.g., normalized adjacency matrix $\NAM$ and transition matrix $\PM$) and node attribute matrix, e.g., $\boldsymbol{\Phi}^{(t)}=p_t(\tilde{\AM})\cdot\XM$, where $p_t(\cdot)$ denotes a polynomial's $t$-order term \cite{sun2023feature,wu2019simplifying,zhu2021interpreting}. As pinpointed in \cite{sun2023feature}, as $i\in \mathbb{Z}$ increases, the feature subspace $\boldsymbol{\Phi}^{(t+i)}$ will be {\em linearly correlated} with $\boldsymbol{\Phi}^{(t)}$, i.e., there exist a weight matrix $\WM$ such that $\|\boldsymbol{\Phi}^{(t)}\WM-\boldsymbol{\Phi}^{(t+i)}\|_2\rightarrow 0$.
Recall that in standard GNNs, all feature subspaces usually share common parameter weights.
For example, given two linearly correlated feature subspaces, $\boldsymbol{\Phi}^{(x)}$ and $\boldsymbol{\Phi}^{(y)}$, the output of a GNN $\mathbf{C}\in \mathbb{R}^{n\times c}$ (e.g., node-class predictions and $c$ is the number of classes) is expressed by $\mathbf{C}=(\lambda_x\boldsymbol{\Phi}^{(x)}+\lambda_y\boldsymbol{\Phi}^{(y)})\cdot \LWM_{c}$, where $\LWM_{c}$ is a transformation weight matrix. However, \cite{sun2023feature} proved that $\mathbf{C}$ can be solely represented by either $\boldsymbol{\Phi}^{(x)}\LWM_{x}$ or $\boldsymbol{\Phi}^{(y)}\LWM_{y}$, indicating that standard GNN models have redundancy and limited expressiveness of the feature space.

By concatenating the original feature subspaces $\boldsymbol{\Phi}^{(t)}$ with adjacency matrix $\AM$, Theorem \ref{lem:feat-space} shows that the GNN models can be more accurate in the recovery of the ground-truth $\mathbf{C}_{\textnormal{exact}} \in \mathbb{R}^{n\times c}$ with learned weight $\LWM^{\prime}_{c}$, compared to the original feature subspaces $\boldsymbol{\Phi}^{(t)}$ with learned weight $\LWM_{c}$.
\begin{theorem}\label{lem:feat-space}
Suppose that the dimensionality $d$ of node attributes $\XM$ satisfies $d\ll n$. The weight matrices $\LWM^{\prime}_{c}$ and $\LWM_{c}$ are the solutions to linear systems $(\boldsymbol{\Phi}^{(t)} \mathbin\Vert \AM)\cdot \LWM^{\prime}_{c}= \mathbf{C}_{\textnormal{exact}}$ and $\boldsymbol{\Phi}^{(t)}\LWM_{c}= \mathbf{C}_{\textnormal{exact}}$, respectively. Then,
\begin{equation*}
\|(\boldsymbol{\Phi}^{(t)} \mathbin\Vert \AM)\cdot \LWM^{\prime}_{c}- \mathbf{C}_{\textnormal{exact}}\|_F < \|\boldsymbol{\Phi}^{(t)}\LWM_{c}- \mathbf{C}_{\textnormal{exact}}\|_F.
\end{equation*}
\begin{proof}
Given any node feature matrix $\boldsymbol{\Psi}\in \mathbb{R}^{n \times \psi}$, the goal of linear regression problem $\boldsymbol{\Psi}\boldsymbol{\Omega}=\mathbf{C}_{\textnormal{exact}}\in \mathbb{R}^{n\times c}$ is to find a weight matrix $\boldsymbol{\Omega} \in \mathbb{R}^{\psi\times c}$ such that $\boldsymbol{\Psi}\boldsymbol{\Omega}$ approximates $\mathbf{C}_{\textnormal{exact}}$ with minimal error. Let $\UM_{\Psi}\boldsymbol{\Sigma}_{\Psi}\VM_{\Psi}^{\top}$ be the exact full singular value decomposition (SVD) of $\boldsymbol{\Psi}$. 
Since the inverse of $\UM_{\Psi}\boldsymbol{\Sigma}_{\Psi}\VM_{\Psi}^{\top}$ is $\VM_{\Psi}\boldsymbol{\Sigma}^{-1}_{\Psi}\UM_{\Psi}^{\top}$ and $\UM_{\Psi}^{\top}\UM_{\Psi}=\VM_{\Psi}^{\top}\VM_{\Psi}=\IM$, we have
\begin{align*}
\UM_{\Psi}\boldsymbol{\Sigma}_{\Psi}\VM_{\Psi}^{\top}\boldsymbol{\Omega} &= \mathbf{C}_{\textnormal{exact}} \\
(\UM_{\Psi}\boldsymbol{\Sigma}_{\Psi}\VM_{\Psi}^{\top})^{-1}\UM_{\Psi}\boldsymbol{\Sigma}_{\Psi}\VM_{\Psi}^{\top}\boldsymbol{\Omega} &= (\UM_{\Psi}\boldsymbol{\Sigma}_{\Psi}\VM_{\Psi}^{\top})^{-1}\mathbf{C}_{\textnormal{exact}} \\
\boldsymbol{\Omega} & = \VM_{\Psi}\boldsymbol{\Sigma}^{-1}_{\Psi}\UM_{\Psi}^{\top}\mathbf{C}_{\textnormal{exact}}.
\end{align*}
which implies that when weight matrix $\boldsymbol{\Omega}$ is $\VM_{\Psi}\boldsymbol{\Sigma}^{-1}_{\Psi}\UM_{\Psi}^{\top}\mathbf{C}_{\textnormal{exact}}$, the best approximation $\mathbf{C}^{\prime}_{\textnormal{exact}}$ of $\mathbf{C}_{\textnormal{exact}}$ is
\begin{align*}
\mathbf{C}^{\prime}_{\textnormal{exact}}=\UM_{\Psi}\boldsymbol{\Sigma}_{\Psi}\VM_{\Psi}^{\top}\VM_{\Psi}\boldsymbol{\Sigma}^{-1}_{\Psi}\UM_{\Psi}^{\top}\mathbf{C}_{\textnormal{exact}}=\UM_{\Psi}\UM_{\Psi}^{\top}\mathbf{C}_{\textnormal{exact}}.
\end{align*}
That is to say, if $\UM_{\Psi}\UM_{\Psi}^{\top}$ is closer to the identity matrix, $\mathbf{C}^{\prime}_{\textnormal{exact}}$ is more accurate. Next, we bound the difference between $\UM_{\Psi}\UM_{\Psi}^{\top}$ and identity matrix $\IM$ as follows:
\begin{align*}
\|\UM_{\Psi}\UM_{\Psi}^{\top}-\IM\|_F&= trace( (\UM_{\Psi}\UM_{\Psi}^{\top}-\IM)^{\top}\cdot (\UM_{\Psi}\UM_{\Psi}^{\top}-\IM))\\
& = trace(\UM_{\Psi}\UM_{\Psi}^{\top} + \IM -2\UM_{\Psi}\UM_{\Psi}^{\top}) \\
& = trace(\IM) - trace(\UM_{\Psi}\UM_{\Psi}^{\top})\\
& = trace(\IM) - trace(\UM_{\Psi}^{\top}\UM_{\Psi})= n-\psi
\end{align*}
By the sub-multiplicative property of matrix Frobenius norm,
\begin{align}
\|\mathbf{C}^{\prime}_{\textnormal{exact}}-\mathbf{C}_{\textnormal{exact}}\|&=\|(\UM_{\Psi}\UM_{\Psi}^{\top}-\IM)\cdot \mathbf{C}_{\textnormal{exact}}\|_F\notag \\
& \le \|\UM_{\Psi}\UM_{\Psi}^{\top}-\IM\|_F\cdot \|\mathbf{C}_{\textnormal{exact}}\|_F \notag \\
&= (n-\psi) \cdot \|\mathbf{C}_{\textnormal{exact}}\|_F.  \label{eq:-uuc}
\end{align}

When $\boldsymbol{\Psi}=\boldsymbol{\Phi}^{(t)}$, we have $\psi=d \ll n$ and $\boldsymbol{\Psi}$ is a thin matrix. As shown in the proof of Theorem 4.2 in \cite{sun2023feature}, $\UM_{\Psi}\UM_{\Psi}^{\top}$ will be rather dense (far from the identity matrix), and hence, rendering $\mathbf{C}^{\prime}_{\textnormal{exact}}$ inaccurate. Also, by Eq. \eqref{eq:-uuc}, the approximation error is $(n-d)\cdot \|\mathbf{C}_{\textnormal{exact}}\|_F$, which is large since $n-d$ is large.

By contrast, when $\boldsymbol{\Psi}=\boldsymbol{\Phi}^{(t)} \mathbin\Vert \AM$, $\psi=n+d$ and we have
\begin{align*}
\boldsymbol{\Psi} & = \UM_{\Phi}\boldsymbol{\Sigma}_{\Phi}\VM_{\Phi}^{\top}  \mathbin\Vert \UM_{A}\boldsymbol{\Sigma}_{A}\VM_{A}^{\top}\\
& = (\UM_{\Phi} \mathbin\Vert \UM_{A})\cdot \begin{pmatrix}
\boldsymbol{\Sigma}_{\Phi}
  & \rvline & 0 \\
\hline
  0 & \rvline &
\boldsymbol{\Sigma}_{A}
\end{pmatrix} \cdot \begin{pmatrix}
\VM_{\Phi}
  & \rvline & 0 \\
\hline
  0 & \rvline &
\VM_{A}
\end{pmatrix}^{\top}
\end{align*}
where $\UM_{\Phi}\boldsymbol{\Sigma}_{\Phi}\VM_{\Phi}^{\top}$ and $\UM_{A}\boldsymbol{\Sigma}_{A}\VM_{A}^{\top}$ are the exact full SVDs of $\boldsymbol{\Phi}^{(t)}$ and $\AM$, respectively.
Similarly, we can derive that
\begin{align*}
\mathbf{C}^{\prime}_{\textnormal{exact}}& =(\UM_{\Phi} \mathbin\Vert \UM_{A})\cdot \begin{pmatrix}
\boldsymbol{\Sigma}_{\Phi}
  & \rvline & 0 \\
\hline
  0 & \rvline &
\boldsymbol{\Sigma}_{A}
\end{pmatrix} \cdot \begin{pmatrix}
\VM_{\Phi}^{\top}
  & \rvline & 0 \\
\hline
  0 & \rvline &
\VM_{A}^{\top}
\end{pmatrix}\\
& \cdot \begin{pmatrix}
\VM_{\Phi}
  & \rvline & 0 \\
\hline
  0 & \rvline &
\VM_{A}
\end{pmatrix} \begin{pmatrix}
\boldsymbol{\Sigma}_{\Phi}
  & \rvline & 0 \\
\hline
  0 & \rvline &
\boldsymbol{\Sigma}_{A}
\end{pmatrix}^{-1} \cdot (\UM_{\Phi} \mathbin\Vert \UM_{A})^{\top} \mathbf{C}_{\textnormal{exact}} \\
& = (\UM_{\Phi} \mathbin\Vert \UM_{A})\cdot (\UM_{\Phi} \mathbin\Vert \UM_{A})^{\top} \mathbf{C}_{\textnormal{exact}} = (\UM_{\Phi}\UM_{\Phi}^{\top} + \UM_{A}\UM_{A}^{\top})\cdot \mathbf{C}_{\textnormal{exact}}.
\end{align*}
According to Theorem 4.1 in \cite{zhang2018arbitrary} and the fact that $\AM$ is a symmetric non-negative matrix, the left singular vectors $\UM_{A}$ of $\AM$ are also the eigenvectors of $\AM$, which are orthogonal and hence $\UM^{-1}_{A}=\UM^{\top}_{A}$. Therefore, we obtain $\UM_{A}\UM_{A}^{\top}=\UM_{A}\UM_{A}^{-1}=\IM$ and
\begin{equation*}
\mathbf{C}^{\prime}_{\textnormal{exact}} = \UM_{\Phi}\UM_{\Phi}^{\top}\mathbf{C}_{\textnormal{exact}} + \mathbf{C}_{\textnormal{exact}}.
\end{equation*}
Hence, by the sub-multiplicative property of matrix Frobenius norm and the relation between Frobenius norm and matrix trace,
\begin{align}
\| \mathbf{C}^{\prime}_{\textnormal{exact}} - \mathbf{C}_{\textnormal{exact}}\|_F & = \| \UM_{\Phi}\UM_{\Phi}^{\top}\mathbf{C}_{\textnormal{exact}} + \mathbf{C}_{\textnormal{exact}} - \mathbf{C}_{\textnormal{exact}} \|_F \notag \\
& = \| \UM_{\Phi}\UM_{\Phi}^{\top}\mathbf{C}_{\textnormal{exact}} \|_F \le \|\UM_{\Phi}\UM_{\Phi}^{\top}\|_F\cdot \|\mathbf{C}_{\textnormal{exact}} \|_F\notag \\
& = trace(\UM_{\Phi}\UM_{\Phi}^{\top})\cdot \|\mathbf{C}_{\textnormal{exact}} \|_F\notag\\
& = trace(\UM_{\Phi}^{\top}\UM_{\Phi})\cdot \|\mathbf{C}_{\textnormal{exact}} \|_F = d\cdot \|\mathbf{C}_{\textnormal{exact}} \|_F,\notag
\end{align}
which is much smaller than $(n-d)\cdot \|\mathbf{C}_{\textnormal{exact}} \|_F$ since $d\ll n$ and completes the proof.
\end{proof}
\end{theorem}

Furthermore, we show that by padding the adjacency matrix as additional node features, we can inject information of high-order (a.k.a. multi-scale or multi-hop) proximity between nodes into the node representations. As revealed in \cite{qiu2018network,tsitsulin2021frede}, network embedding methods \cite{yin2019scalable,tsitsulin2018verse,zhang2018arbitrary,yang2020homogeneous,grover2016node2vec} achieve high effectiveness through implicitly or explicitly factorizing the high-order proximity matrix of nodes 
\begin{equation}\label{eq:NE}
\sum_{i=0}^{t}{w_i\NAM^{i}}\ \text{or}\ \sum_{i=0}^{t}{w_i\PM^{i}}
\end{equation}
or its element-wise logarithm.
If we substitute $\XM \mathbin\Vert \AM$ for the original attribute matrix $\XM$ in Eq.~\eqref{eq:poly}, the node representations at $(t+1)$-layer in GNNs can be formulated as
\begin{align*}
\HM^{(t)} &= f_{\textnormal{poly}}(\NAM,t)\cdot (\XM \mathbin\Vert \AM)\cdot\LWM = \left(f_{\textnormal{poly}}(\NAM,t)\XM \mathbin\Vert f_{\textnormal{poly}}(\NAM,t)\AM\right)\cdot \LWM.
\end{align*}
Note that $f_{\textnormal{poly}}(\NAM,t)\cdot\AM=f_{\textnormal{poly}}(\NAM,t)\cdot\DM^{1/2}\NAM\DM^{1/2}$, which can be rewritten as the form similar to Eq. \eqref{eq:NE} by properly choosing weight $w_i$ for $0\le i\le t$. Similar result can be derived when $\NAM$ is replaced by $\PM=\DM^{-1}\AM$. Hence, $\HM^{(t)}$ contains the information of the high-order proximity between nodes.
Recently, researchers validate the effectiveness of using the high-order node proximity as node features in improving the expressiveness of GNNs ~\cite{Velingker2022AffinityAwareGN}.



\eat{
\stitle{Expressive Power by 1-WL Test}
Lemma 2 in \cite{xu2018powerful} and Theorem 1 in \cite{morris2019weisfeiler} show that MP-GNNs are at most as powerful as the 1-WL test in distinguishing different graph isomorphism classes, indicating the representation power of MP-GNNs is upper-bounded by the 1-WL test. With the example in Figure \ref{fig:WLtest}, Theorem \ref{lem:1-WL} theoretically proves that using the adjacency matrix as additional node features enhances the expressive power of MP-GNNs.
\vspace{-2ex}
\begin{figure}[!h]
    \centering
    \includegraphics[width=0.7\columnwidth]{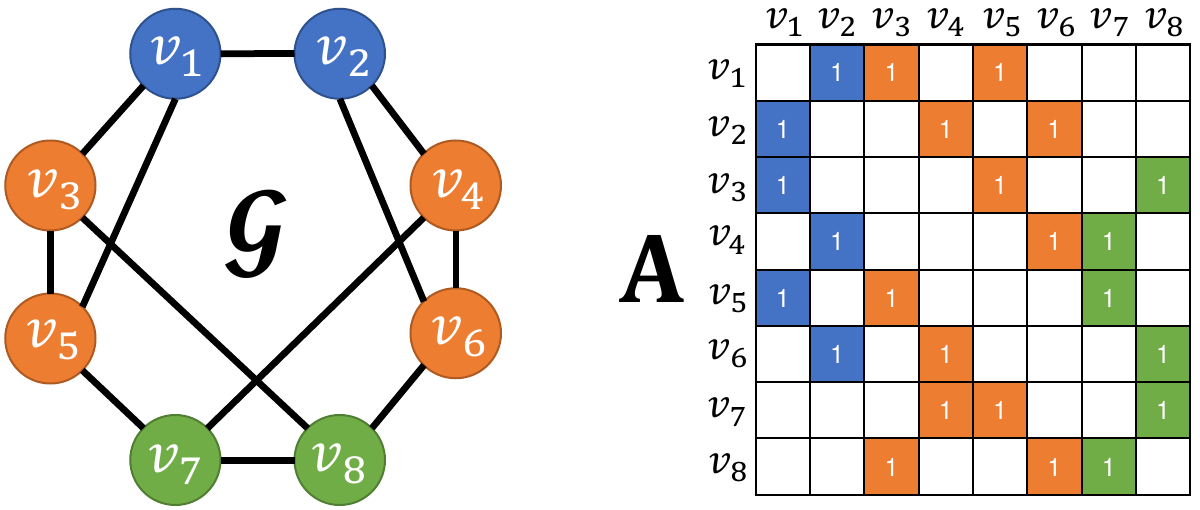}
    \vspace{-2mm}
    \caption{An Example Graph with 3 Isomorphism Classes}\label{fig:WLtest}
    \vspace{-2mm}
\end{figure}
\begin{theorem}\label{lem:1-WL}
MP-GNNs with adjacency matrix $\AM$ as additional node features are strictly more powerful than the WL-1 test.
\end{theorem}
}

\subsection{Algorithmic Details and Analysis}\label{sec:module-1-add}

The pseudo-code of our hybrid method for computing the sketched adjacency matrix $\AM^{\prime}$ is outlined in Algorithm \ref{alg:sketch}.

\begin{algorithm}[h]
\caption{Sketched Adjacency Matrix Construction}\label{alg:sketch}
\KwIn{Graph $\G=(\V,\EDG)$, dimension $k$, the number $T$ of iterations, weight $\beta$, and the size $|\C|$ of centroids}
\KwOut{The sketched adjacency matrix $\AM^{\prime}$}
Construct count-sketch matrix $\RM$ as per Eq. \eqref{eq:sketch-A}\;
Select a set $\C$ of nodes from $\V$ with top-$|\C|$ in-degrees\;
Initialize an $n\times |\C|$ matrix $\boldsymbol{\Pi}^{(0)}$\;
\lFor{$v_i\in \C$}{
$\boldsymbol{\Pi}^{(0)}_{i,i}\gets 1$
}
$\boldsymbol{\Pi}\gets (1-\alpha)\cdot\boldsymbol{\Pi}^{(0)}$\;
\For{$t\gets 1$ to $T$}{
$\boldsymbol{\Pi} \gets \alpha\cdot \PM \boldsymbol{\Pi} + (1-\alpha)\cdot\boldsymbol{\Pi}^{(0)}$\;
}
\lFor{$v_i\in \C$}{
$\pi(v_i)\gets \sum_{v_j\in \V}{\boldsymbol{\Pi}_{j,i}}$
}
Sort nodes $v_i\in \C$ in descending order according to $\pi(v_i)$\;
Let $\C_k$ be the top-$k$ nodes in $\C$\;
$\SM\gets \mathbf{0}^{k\times n}$\;
\For{$v_i\in \V$}{
$v_{j^{\ast}} \gets \max_{v_j\in \C_k}{\boldsymbol{\Pi}_{i,j}}$\;
$\SM_{j^{\ast},i}\gets 1$\;
}
Apply a row-wise $L_2$ normalization over $\SM$\;
Compute $\AM^{\prime}$ by Eq.~\eqref{eq:A-prime}\;
\Return{$\AM^{\prime}$}\;
\end{algorithm}

\begin{algorithm}[!h]
\caption{Sparsified Adjacency Matrix Construction}\label{alg:sparsify}
\KwIn{The adjacency matrix $\AM$, node representations $\HM^{(0)}$, sparsification ratio $\rho$}
\KwOut{The sparsified adjacency matrix $\AM^{\circ}$}
\For{$e_{i,j}\in \EDG$}{
Compute edge weight $w(e_{i,j})$ by Eq. \eqref{eq:e-weight}\;
$\AM_{i,j}\gets w(e_{i,j}),\ \AM_{j,i}\gets w(e_{i,j})$\;
}
\For{$v_i\in \V$}{
Compute $d_w(v_i)$ by Eq. \eqref{eq:d-weight}\;
}
\For{$e_{i,j}\in \EDG$}{
Compute the centrality $C_w(e_{i,j})$ of edge $e_{i,j}$ by Eq. \eqref{eq:e-centrality}\;
}
Let $\EDG_{\rho}$ be the edges with $m\cdot\rho$ smallest centrality values\;
\For{$e_{i,j}\in \EDG_{\rho}$}{
$\AM_{i,j}\gets 0,\ \AM_{j,i}\gets 0$
}
\Return{$\AM^{\circ}\gets \AM$}\;
\end{algorithm}

\begin{table*}[!t]
\centering
\renewcommand{\arraystretch}{0.9}
\begin{small}
\caption{Complete Statistics of Datasets.}\label{tbl:exp-data-full}
\vspace{-3ex}
\resizebox{\textwidth}{!}{%
\begin{tabular}{l|r|r|r|c|c|c|c}
	\hline
	{\bf Dataset} & \multicolumn{1}{c|}{\bf \#Nodes } & \multicolumn{1}{c|}{\bf \#Edges } & \multicolumn{1}{c|}{\bf \#Attr. } & \multicolumn{1}{c|}{\bf \#Classes} & \multicolumn{1}{c|}{\bf Avg. Degree} & \multicolumn{1}{c|}{\bf Homophily Ratio} & \multicolumn{1}{c|}{\bf Traing/Val./Test} \\
	\hline
	\hline
    {\em Photo} \cite{shchur2018pitfalls} & 7,650 & 238,162 & 745 & 8 & 31.1 (HDG) & 0.83 (homophilic) & 60\%/20\%/20\%  \\
    {\em WikiCS} \cite{mernyei2020wiki} & 11,701 &  431,726 & 300 & 10 & 36.9 (HDG) & 0.65 (homophilic) & 60\%/20\%/20\%  \\
    {\em Reddit2} \cite{zeng2019graphsaint} & 232,965 & 23,213,838 & 602 & 41 & 99.6 (HDG) & 0.78 (homophilic) & 66\%/10\%/24\% \\
    {\em Amazon2M} \cite{chiang2019cluster} & 2,449,029 & 61,859,140 & 100 & 47 & 25.3 (HDG) & 0.81 (homophilic) & 8\%\%2\%90\% \\
    \hline
    {\em Squirrel} \cite{pei2019geom} & 5,201 &  396,846 & 2,089 & 5 & 76.3 (HDG) & 0.22 (heterophilic) & 60\%/20\%/20\% \\
    {\em Penn94} \cite{hu2020open} & 41,554 &  1,362,229 & 128 & 2 & 32.8 (HDG) & 0.47 (heterophilic) & 50\%/25\%/20\% \\
    {\em Ogbn-Proteins} \cite{hu2020open} &  132,534 & 39,561,252 & 8 & 112  & 298.5 (HDG) & 0.38 (heterophilic) & 65\%/16\%/19\% \\
    {\em Pokec}~\cite{lim2021large} & 1,632,803 &  30,622,564 & 65 & 2 & 18.8 (HDG) & 0.45 (heterophilic) & 50\%/25\%/24\% \\
    \hline
    {\em Cora} \cite{yang2016revisiting} & 2,708 & 5,429 & 1,433 & 7 & 2.0 (LDG) & 0.81 (homophilic) &  60\%20\%20\% \\
    {\em arXiv-Year} \cite{hu2020open} & 169,343 &   1,166,243 & 128 & 5 & 6.9 (LDG) & 0.22 (heterophilic) & 50\%/25\%/21\% \\
    \hline
\end{tabular}%
}
\end{small}
\end{table*}

\begin{table*}[!t]
\centering
\renewcommand{\arraystretch}{0.9}
\caption{Hyperparameters of \algo.}\vspace{-3mm}
\begin{footnotesize}
\addtolength{\tabcolsep}{-0.25em}
\begin{tabularx}{\textwidth}{l|X|X|X|X|X|X|X|X}
\hline
\multirow{1}{*}{\bf Method} & \multicolumn{1}{c|}{\bf{ {\em Photo}}} & \multicolumn{1}{c|}{\bf{ {\em WikiCS}}}  & \multicolumn{1}{c|}{\bf{ {\em Reddit2}}}  & \multicolumn{1}{c|}{\bf{ {\em Amazon2M}}}  & \multicolumn{1}{c|}{\bf{ {\em Squirrel}}} & \multicolumn{1}{c|}{\bf{ {\em Penn94}}} & \multicolumn{1}{c|}{\bf{ {\em Ogbn-Proteins}}} & \multicolumn{1}{c}{\bf{ {\em Pokec}}} \\ \cline{2-9}
\hline
\hline
GCN & $k=128,\gamma=0.3, \rho=0.3$, lr$=$0.001, weight-decay$=$1e-5,  dropout$=$0.5& 	$k=128,\gamma=0.3, \rho=$0.1, lr$=$0.05, weight-decay$=$1e-5,  dropout$=$0.3	&	$k=128,\gamma=0.5,\rho=$0.7, lr$=$0.01, weight-decay$=$0.0,  dropout$=$0.5	&	$k=128,\gamma=0.5,\rho=$0.7, lr$=$, weight-decay$=$1e-5,  dropout$=$0.5	&	$k=128,\gamma=1.0,\rho=$0.5, lr$=$0.05, weight-decay$=$1e-5,  dropout$=$0.1	&	$k=1024,\gamma=0.7,\rho=$0.1, lr$=$0.01, weight-decay$=$1e-5,  dropout$=$0.1&	$k=64,\gamma=0.5,\rho=$0.9, lr$=$0.01, weight-decay$=$1e-5,  dropout$=$0.5	&	$k=512,\gamma=0.5,\rho=$0.1, lr$=$0.001, weight-decay$=$1e-5,  dropout$=$0.1	  \\ \hline
GAT &$k=128,\gamma=0.3,\rho=0.8$, lr$=$0.005, weight-decay$=$1e-5,  dropout$=$0.1& $k=128,\gamma=0.3,\rho=$0.1, lr$=$0.01, weight-decay$=$0.0,  dropout$=$0.3		&$k=128,\gamma=0.5,\rho=$0.7, lr$=$0.01, weight-decay$=$1e-5,  dropout$=$0.5&$k=128,\gamma=0.5,\rho=$0.5, lr$=$0.01, weight-decay$=$1e-5,  dropout$=$0.5&$k=128,\gamma=1.0,\rho=$0.5, lr$=$0.05, weight-decay$=$0.0,  dropout$=$0.1	&$k=128,\gamma=1.0,\rho=$0.1, lr$=$0.005, weight-decay$=$1e-5,  dropout$=$0.1&$k=128,\gamma=0.5,\rho=$0.9, lr$=$0.01, weight-decay$=$1e-5,  dropout$=$0.5&$k=256,\gamma=0.5,\rho=$0.1, lr$=$0.001, weight-decay$=$1e-5,  dropout$=$0.1\\  \hline
SGC &$k=128,\gamma=0.7,\rho=$0.2, lr$=$0.005, weight-decay$=$1e-5,  dropout$=$0.1		&$k=128,\gamma=0.3,\rho=$0.2, lr$=$0.001, weight-decay$=$1e-5,  dropout$=$0.3 		&$k=128,\gamma=0.5,\rho=$0.7, lr$=$0.01, weight-decay$=$1e-5,  dropout$=$0.5&$k=128,\gamma=0.5,\rho=$0.5, lr$=$0.01, weight-decay$=$1e-5,  dropout$=$0.5&$k=128,\gamma=1.0,\rho=$0.5, lr$=$0.05, weight-decay$=$0.0,  dropout$=$0.1&$k=128,\gamma=0.5,\rho=$0.1, lr$=$0.05, weight-decay$=$1e-5,  dropout$=$0.1&$k=128,\gamma=0.5,\rho=$0.9, lr$=$0.01, weight-decay$=$1e-5,  dropout$=$0.5&$k=512,\gamma=0.2,\rho=$0.1, lr$=$0.05, weight-decay$=$1e-5,  dropout$=$0.5	\\  \hline
APPNP &$k=128,\gamma=0.2,\rho=$0.3, lr$=$0.005, weight-decay$=$1e-5,  dropout$=$0.1	&$k=128,\gamma=0.3,\rho=$0.2, lr$=$0.05, weight-decay$=$1e-5,  dropout$=$0.3 	&$k=128,\gamma=0.5,\rho=$0.7, lr$=$0.01, weight-decay$=$1e-5,  dropout$=$0.5&$k=128,\gamma=0.5,\rho=$0.5, lr$=$0.01, weight-decay$=$1e-5,  dropout$=$0.5&$k=128,\gamma=1.0,\rho=$0.5, lr$=$0.05, weight-decay$=$0.0,  dropout$=$0.1&$k=128,\gamma=0.5,\rho=$0.1, lr$=$0.05, weight-decay$=$1e-5,  dropout$=$0.1&$k=128,\gamma=0.5,\rho=$0.9, lr$=$0.01, weight-decay$=$1e-5,  dropout$=$0.5&$k=256,\gamma=0.5,\rho=$0.1, lr$=$0.001, weight-decay$=$1e-5,  dropout$=$0.1\\  \hline
GCNII &$k=128,\gamma=0.3,\rho=$0.3, lr$=$0.001, weight-decay$=$1e-5,  dropout$=$0.0	&$k=128,\gamma=0.2,\rho=$0.2, lr$=$0.01, weight-decay$=$5e-5,  dropout$=$0.3 	&$k=128,\gamma=0.5,\rho=$0.7, lr$=$0.01, weight-decay$=$0.0,  dropout$=$0.5&$k=128,\gamma=0.5,\rho=$0.5, lr$=$0.01, weight-decay$=$1e-5,  dropout$=$0.5&$k=128,\gamma=1.0,\rho=$0.5, lr$=$0.05, weight-decay$=$1e-5,  dropout$=$0.1&$k=128,\gamma=0.5,\rho=$0.1, lr$=$0.01, weight-decay$=$1e-5,  dropout$=$0.1&$k=128,\gamma=0.5,\rho=$0.9, lr$=$0.05, weight-decay$=$0.0,  dropout$=$0.5&$k=256,\gamma=0.5,\rho=$0.1, lr$=$0.001, weight-decay$=$1e-5,  dropout$=$0.1 \\  \hline
\end{tabularx}
\end{footnotesize}
\label{tbl:params}
\vspace{-1ex}
\end{table*}

\stitle{Complexity Analysis of Count-Sketch} Recall that $\AM$ is a sparse matrix containing $m$ non-zero entries and $\RM=\boldsymbol{\Phi}\boldsymbol{\Delta}$ where $\boldsymbol{\Delta}$ is a $n\times n$ diagonal matrix and each column in $\boldsymbol{\Phi}\in \mathbb{R}^{k\times n}$ solely has a single non-zero entry. Therefore, the sparse matrix multiplication $\AM\RM^\top$ in Eq. \eqref{eq:sketch-A} (Line 1 in Algorithm \ref{alg:sketch}) consumes $O(m)$ time.

\stitle{Complexity Analysis of RWR-Sketch}
According to Lines 2–15 in Algorithm \ref{alg:sketch}, the sorting at Line 2 takes $O(n + |\C|\cdot\log n)$ time when using the max-heap. Notice that each of $T$ iterations at Lines 6-7 conducts sparse matrix multiplication, where $\PM$ contains $m$ non-zero entries and $\boldsymbol{\Pi}$ is of size $n\times |\C|$. Hence, Lines 4-8 consume $O(m\cdot |\C|)$ time. Line 9 needs a full sorting of set $\C$, which incurs an $O(|\C|\cdot\log(|\C|))$ cost. As for Lines 12-14, for each node, the cost of Lines 13-14 is $O(|\C|)$, resulting in a total cost of $O(n\cdot |\C|)$. The normalization of the $k\times n$ matrix $\SM$ requires $O(n)$ time as each column has only one non-zero element. After constructing $\SM$ wherein each column has only one non-zero entry, the sketching operation $\AM\cdot \SM^{\top}$ is a sparse matrix multiplication, which can be done in $O(m)$ time. Therefore, RWR-Sketch takes $O(m+m\cdot |\C|)$ time in total.

The overall computational cost of Algorithm \ref{alg:sketch} is bounded by $O(m\cdot |\C|)$, which can be reduced to $O(m)$ since $|C|$ can be regarded as a constant.

\section{Module II: Graph Sparsification}\label{sec:sparsify-add}

\stitle{Pseudo-code}
Algorithm \ref{alg:sparsify} displays the pseudo-code of our topology- and attribute-aware graph sparsification.

\stitle{Complexity Analysis}
Since initial node representations $\HM^{(0)}$ are a $n\times h$ matrix, the edge reweighting at Line 2 in Algorithm \ref{alg:sparsify} takes $O(mh)$ time in total. Both the computation of ``degrees'' of nodes at Lines 4-5 and the calculation of edge centrality values at Lines 6-7 require $O(m)$ time. The partial sorting for extracting the edges with $m\cdot\rho$ smallest centrality values at Line 8 consumes $O(m+m\rho\log{m})$ time. Lines 9-10 needs $O(m\rho)$ time for processing all the $m\cdot\rho$ edges. Therefore, the total computational complexity is bounded by $O(mh+m\log{m})$.


%% file: tex/addexp.tex
\section{Additional Experiments}

\subsection{Dataset details}\label{sec:datasets-more}
Table \ref{tbl:exp-data-full} presents the full statistics as well as the details of the train/validation/test splits of the 8 datasets in our experiments. For a graph with node class labels $y_v \in \Y$, we define its {\em homophily ratio} (HR) as the fraction of homophilic edges linking same-class nodes \cite{zhu2020beyond}: $HR = |{(u, v) | (u, v) \in \EDG \wedge y_u = y_v}|/|\EDG|$. 

\noindent\textbf{Photo} \cite{shchur2018pitfalls} is a segment of the Amazon co-purchase graph, where nodes represent goods and edges indicate that two goods are frequently purchased together. The node features are extracted from the product reviews and node class labels correspond to product categories. \\
\textbf{WikiCS}~\cite{mernyei2020wiki} is collected from Wikipedia, where nodes are wiki pages and edges represent hyperlinks between pages. The node attributes are word embeddings constructed from the articles and node class labels correspond to 10 branches of computer science. \\
\textbf{Reddit2}~\cite{zeng2019graphsaint} is constructed based on Reddit posts. The edge between two nodes (i.e., posts) indicates that the same user comments on both of them. Node class labels are the communities where the nodes are from, and the node attributes are off-the-shelf 300-dimensional GloVe CommonCrawl word vectors of the post. \\
\textbf{Amazon2M}~\cite{hu2020open} is an Amazon product co-purchasing network where nodes and edges represent the products and co-purchasing relationships between products, respectively. The node attributes are the bag-of-words of the description of products and node classes represent product categories.\\
\textbf{Squirrel} is a network consisting of Wikipedia pages on ``squirrel'' topics, respectively. Nodes are Wikipedia articles and edges are hyperlinks between to pages. Node attributes of Squirrel are a group of selected noun from the article. Nodes are divided into different classes based on their traffic. \\
\textbf{Penn94}~\cite{hu2020open} is a subgraph extracted from Facebook in which nodes represent students and edges are their friendships. The nodal attributes include major, second major/minor, dorm/house, year, and high school. The node class labels are students' genders.\\
\textbf{Ogbn-Proteins}~\cite{hu2020open} is a protein association network. Nodes represent proteins, and edges are associations between proteins. Edges are multi-dimensional features, where each dimension is the approximate confidence of different association types in the range of $[0,1]$. Each node can carry out multiple functions,and each function represents a label. A multi-label binary classification task on this graph is to predict the functions of each node (protein). \\
\textbf{Pokec}~\cite{lim2021large} is extracted from a Slovak online social network, whose nodes correspond to users and edges represent directed friendships. Node attributes are constructed from users' profiles, such as geographical region and age. The users' genders are taken as node class labels.\\
\textbf{Cora}~\cite{yang2016revisiting} is a citation network where nodes represent papers and node attributes are bag-of-words representations of the paper. \\
\textbf{arXiv-year}~\cite{lim2021large} is also a citation network. Nodes stand for papers, and edges represent the citation relationships between papers. For each node (i.e., paper), its attributes are the Word2vec representations of its title and abstract. Node class labels correspond to the published years of papers.


\subsection{Implementation Deatils and Hyperparameter 
Settings}\label{sec:baseline-more}
\eat{
GCN \cite{kipf2016semi}: https://github.com/tkipf/gcn\\
GAT \cite{velivckovic2018graph}: https://github.com/gordicaleksa/pytorch-GAT\\
SGC \cite{wu2019simplifying}: https://github.com/Tiiiger/SGC\\
APPNP \cite{gasteiger2018predict}: https://github.com/benedekrozemberczki/APPNP\\
GCNII \cite{chen2020simple}: https://github.com/chennnM/GCNII\\
\renchi{Need to describe the implementation of the baselines and \algo. implemented by PyTorch Geometric \cite{Fey/Lenssen/2019}?}
}

We implement the 5 baseline GNN models and \algo based on PyTorch Geometric~\cite{Fey/Lenssen/2019}. We collect homophilic graphs {\em Cora}, {\em Photo}, {\em WikiCS} from \cite{cui2023mgnn}\footnote{\url{https://github.com/GuanyuCui/MGNN}}, heterophilic graphs {\em Squirrel}, {\em Penn94} and {\em Pokec} from ~\cite{lim2021large}\footnote{\url{https://github.com/CUAI/Non-Homophily-Large-Scale}}. Large datasets {\em Reddit2}, {\em Ogbn-Proteins} and {\em Amazon2M} are downloaded from DSpar~\cite{liu2023dspar}\footnote{\url{https://github.com/warai-0toko/DSpar_tmlr}}.

For all the tested GNN models, we set the number of layers to 2, and hidden dimension is in  $\{128,256 \}$. The weight decay is in the range from $5\times10^{-5}$ to 0, the learning rate is in the interval $[10^{-3}, 5\times10^{-2}]$, and the dropout rate is in $\{ 0.1, 0.2, 0.3, 0.4, 0.5\}$. We set the total number of training epochs in the range \{800, 1000, 1500, 1800\}. For \algo, we set the number of pretraining epochs ($n_p$ in Module I of \algo) to 128. We set the random walk steps $T$ to 2, and the weight of
RWR-Sketch $\beta$ to 1. 

Table \ref{tbl:params} reports the settings of parameters used in \algo when working in tandem with GNN models: GCN, GAT, SGC, APPNP, and GCNII on 8 experimented datasets. 

\begin{table}[!h]
\centering
\renewcommand{\arraystretch}{0.9}
\caption{Node classification results on LDGs.}\vspace{-3mm}
\begin{small}
\addtolength{\tabcolsep}{-0.25em}
\begin{tabular}{c|c| c}
\hline
\multirow{2}{*}{\bf Method} & \multicolumn{1}{c|}{\bf{ {\em Cora }}} & \multicolumn{1}{c}{\bf{ {\em arXiv-Year}}}  \\ \cline{2-3}
& Acc (\%)  & Acc (\%)  \\ 
\hline
\hline
GCN    & {\bf 87.95±0.35} & 			44.84±0.38		\\
GCN + \algo &86.18±1.03& 			{\bf 46.84±0.18}	 		\\ \hline
GAT    & {\bf 89.19±0.51}	& 			{\bf 43.55±0.2}	\\ 
GAT + \algo &88.75±0.50& 			{ 42.65±0.28}			\\ \hline
SGC    &85.90±0.47& 			38.9±0.19			\\ 
SGC + \algo & {\bf 88.71±0.39} & 			{\bf 41.56±0.24}	 		\\ \hline
APPNP    & {\bf 88.71±0.44} & 			40.22±0.26		\\ 
APPNP + \algo &88.38 0.95& 			{\bf 41.54±0.15}			\\ \hline
GCNII    & {\bf 86.46±0.29} & 			46.39±0.2			\\ 
GCNII + \algo &86.22±1.48& 			{\bf 50.48±0.17}		\\
\hline
\end{tabular}
\end{small}
\label{tbl:LDGs}
\vspace{0ex}
\end{table}

\subsection{Performance of \algo on Low-degree Graphs (LDGs)}\label{sec:exp-LDGs}

Table \ref{tbl:LDGs} shows the node classification results of five GNN models and their \algo-augmented counterparts on two low-degree graphs {\em Cora} and {\em arXiv-Year}. Particularly, on {\em Cora} with average node degree 2.0, we can observe that \algo slightly degrade the classification performance of most GNN backbones except SGC. In contrast, on graph {\em arXiv-Year} with higher average degree (6.9), \algo can promote the classification accuracy of four GNN models (GCN, SGC, APPNP, and GCNII) with remarkable gains and lead to performance degradation for GAT. The observations indicate that \algo is more suitable for GNNs over HDGs as it will cause information loss and curtail the classification performance of GNNs on graphs with scarce connections.

